\documentclass[twoside]{article}

\usepackage[accepted]{aistats2022}
\usepackage[round]{natbib}

\usepackage[utf8]{inputenc} 
\usepackage[T1]{fontenc}    
\usepackage{hyperref}       
\usepackage{url}            
\usepackage{booktabs}       
\usepackage{amsfonts}       
\usepackage{nicefrac}       
\usepackage{microtype}      
\usepackage{amsmath, amssymb, amsthm}
\usepackage{xcolor}
\usepackage{listings}
\usepackage{graphicx}
\usepackage{subfiles}
\usepackage{multirow}
\usepackage{wrapfig}
\usepackage{dsfont}
\usepackage{subcaption}
\usepackage[rightcaption]{sidecap}
\usepackage{placeins}
\usepackage{xfrac}
\usepackage[abs]{overpic}
\usepackage{pict2e}

\definecolor{codegreen}{rgb}{0,0.6,0}
\definecolor{codegray}{rgb}{0.5,0.5,0.5}
\definecolor{codepurple}{rgb}{0.58,0,0.82}
\definecolor{backcolour}{rgb}{0.95,0.95,0.92}
\definecolor{summarygray}{rgb}{0.5, 0.5, 0.5}

\lstdefinestyle{mystyle}{
    backgroundcolor=\color{backcolour},   
    commentstyle=\color{codegreen},
    keywordstyle=\color{magenta},
    numberstyle=\tiny\color{codegray},
    stringstyle=\color{codepurple},
    basicstyle=\ttfamily\footnotesize,
    breakatwhitespace=false,         
    breaklines=true,                 
    captionpos=b,                    
    keepspaces=true,                 
    numbers=left,                    
    numbersep=5pt,                  
    showspaces=false,                
    showtabs=false,                  
    tabsize=2
}
\newtheorem{proposition}{Proposition}[section]
\newtheorem{definition}{Definition}[section]

\DeclareMathOperator*{\argmin}{\arg\!\min}

\newcommand{\bs}[1]{\boldsymbol{#1}} 
\renewcommand{\b}[1]{\mathbf{#1}}	
\renewcommand{\vec}[1]{\mathbf{#1}}
\newcommand{\mat}[1]{\mathbf{#1}}
\newcommand{\dif}[0]{\mathrm{d}} 

\newcommand{\Hess}{\text{Hess}}
\newcommand{\KL}{\mathrm{KL}}

\newcommand{\x}{\vec{x}}
\newcommand{\z}{\vec{z}}

\def\T{^{\intercal}} 
\def\R{\mathbb{R}}  
 
\newcommand{\M}{\mathcal{M}} 
\newcommand{\Z}{\mathcal{Z}} 
\newcommand{\X}{\mathcal{X}} 
\newcommand{\HH}{\mathcal{H}} 
\newcommand{\I}{\mathbb{I}} 
\newcommand{\inner}[2]{\langle#1,#2\rangle}
\newcommand{\vectorize}[1]{\mathrm{vec}[#1]}
\newcommand{\parder}[2]{\frac{\partial #1}{\partial #2}}

\usepackage{smartdiagram}
\hypersetup{
    colorlinks,
    linkcolor={red!50!black},
    citecolor={blue!50!black},
    urlcolor={blue!80!black}
}

\begin{document}

%

%
\runningauthor{Georgios Arvanitidis, Miguel González-Duque, Alison Pouplin, Dimitris Kalatzis, S{\o}ren Hauberg}

\twocolumn[

\aistatstitle{Pulling back information geometry}

\aistatsauthor{Georgios Arvanitidis\textsuperscript{{\normalfont * 1 2}} \And Miguel González-Duque\textsuperscript{{\normalfont * 3}} \And Alison Pouplin\textsuperscript{{\normalfont * 1}}}
\aistatsauthor{Dimitris Kalatzis\textsuperscript{{\normalfont * 1}} \And S{\o}ren Hauberg\textsuperscript{{\normalfont * 1}}}

\aistatsaddress{\textsuperscript{{\normalfont 1}} Technical University of Denmark, Section for Cognitive Systems, Lyngby, Denmark \\ \textsuperscript{{\normalfont 2}} Max Planck Institute for Intelligent Systems, T{\"u}bingen, Germany \\  \textsuperscript{{\normalfont 3}} IT University of Copenhagen, Creative AI Lab, Copenhagen, Denmark} ]

\begin{abstract}
  Latent space geometry has shown itself to provide a rich and rigorous framework for interacting with the latent variables of deep generative models. 
  The existing theory, however, relies on the decoder being a Gaussian distribution as its simple reparametrization allows us to interpret the generating process as a random projection of a deterministic manifold.
  Consequently, this approach breaks down when applied to decoders that are not as easily reparametrized.
  We here propose to use the Fisher-Rao metric associated with the space of decoder distributions as a reference metric,  which we pull back to the latent space.
  We show that we can achieve meaningful latent geometries for a wide range of decoder distributions for which the previous theory was not applicable, opening the door to `black box' latent geometries.
\end{abstract}

\section{Introduction}
\label{sec:intro}

Generative models such as \emph{variational autoencoders (VAEs)} \citep{kingma:iclr:2014, rezende:icml:2014} and \emph{generative adversarial networks (GANs)} \citep{goodfellow:neurips:2014} provide state-of-the-art density estimators for high dimensional data. The underlying assumption is that data $\x \in \X$ lie near a low-dimensional manifold $\M \subset \X$, which is parametrized through a low-dimensional \emph{latent representation} $\z \in \Z$. As data is finite and noisy, we only recover a probabilistic estimate of the true manifold, which, in VAEs, is represented through a decoder distribution $p(\x | \z)$. Our target is the geometry of this \emph{random manifold}.

The geometry of the manifold has been shown to carry great value when systematically interacting with the latent representations, as it provides a stringent solution to the \emph{identifiability problem} that plagues latent variable models \citep{tosi:uai:2014, arvanitidis:iclr:2018, hauberg:only:2018}. For example, this geometry has allowed VAEs to discover latent evolutionary signals in proteins \citep{detlefsens:proteins:2020}, provide efficient robot controls \citep{scannellTrajectory2021, chen2018active, hadi:rss:2021}, improve latent clustering abilities \citep{yang:arxiv:2018, arvanitidis:iclr:2018} and more. The fundamental issue with these geometric approaches is that the studied manifold is inherently a stochastic object, but classic differential geometry only supports the study of \emph{deterministic} manifolds. To bridge the gap, \citet{eklund:arxiv:2019} have shown how VAEs with a Gaussian decoder family can be viewed as a random projection of a deterministic manifold, thereby making the classic theories applicable to the random manifold.

A key strength of VAEs is that they can model data from diverse modalities through the choice of decoder distribution $p(\x | \z)$. For discrete data, we use categorical decoders, while for continuous data we may opt for a Gaussian, a Gamma or whichever distribution best suits the data. However, for non-Gaussian decoders, there exists no useful approach for treating the associated random manifold as deterministic, which prevents us from systematically interacting with the latent representations without being subjected to identifiability issues. This limitation motivates the current work.

\textbf{In this paper}, we provide a general framework that allows us to interact with the geometry of almost any random manifold. The key, and simple idea is to reinterpret the decoder as spanning a deterministic manifold in the space of probability distributions $\HH$, rather than a random manifold in the observation space (see Fig.~\ref{fig:teaser_img}). Calling on classical \emph{information geometry} \citep{amari:2016, nielsen:2020}, we show that the learned manifold is a Riemannian manifold of $\HH$, and provide the corresponding computational tools. The approach is applicable to any family of decoders for which the KL-divergence can be differentiated, allowing us to work with a wide range of models from a single codebase.\looseness=-1

\section{The geometry of generative models}
\label{sec:riemannian_geometry}

As a starting point, consider the deterministic generative model given by a prior $p(\z)$ and a decoder $f: \Z=\R^d \rightarrow \X=\R^D$, which is assumed to be a smooth immersion. The latent representation $\z$ of an observation $\x$ is generally not \emph{identifiable}, meaning that one can recover different latent representations that give rise to equally good density estimates. For example, let $g: \Z \rightarrow \Z$ be a smooth invertible function such that $\z \sim p(\z) \Leftrightarrow g(\z) \sim p(\z)$, then the latent representation $g(\z)$ coupled with the decoder $f \circ g^{-1}$ gives the same density estimate as $\z$ coupled with $f$ \citep{hauberg:only:2018}. Practically speaking, the identifiability issue implies that it is improper to view the latent space $\Z$ as being Euclidean, as any reasonable view of $\Z$ should be invariant to reparametrizations $g$.
  
  The classic geometric solution to the identifiability problem is to define any quantity of interest in the observation space $\X$ rather than the latent space $\Z$. For example, the length of a curve $\gamma: [0, 1] \rightarrow \Z$ in the latent space can be defined as its length measured in $\X$ on the manifold $\M=f(\Z)$ with $N \rightarrow +\infty$ as:
  \begin{align}
    \mathrm{L}(\gamma)
      &=\sum_{n=1}^{N-1} \| f(\gamma(t_{n+1}))\!-\!f(\gamma(t_{n})) \| = \int_0^1 \|  \dot{f}(\gamma(t)) \| \dif t \nonumber\\
      &= \int_0^1 \sqrt{ \dot{\gamma}(t)\T \mat{J}_f({\gamma(t)})\T \mat{J}_f({\gamma(t)}) \dot{\gamma}(t) } \dif t,
     \label{eq:curve_length}
  \end{align}
  where $t_n = \sfrac{n}{N}$ and $t_{n+1} = \sfrac{n+1}{N}$ and we used the chain rule $\partial_t f(\gamma(t)) = \b{J}_f(\gamma(t))\dot{\gamma}(t)$ with $\dot{\gamma}(t) = \partial_t \gamma(t)$ being the curve derivative, and $\mat{J}_f({\gamma(t)})\in\R^{D\times d}$ the Jacobian of $f$ at $\gamma(t)$. This construction shows how we may calculate lengths in the latent space with respect to the metric of the observation space, which is typically assumed to be the Euclidean, but other options exist \citep{arvanitidis:aistats:2021}. In this way, the symmetric positive definite matrix $\mat{J}_f({\gamma(t)})\T \mat{J}_f({\gamma(t)})$ is denoted by $\b{M}({\gamma(t)})\in\R_{\succ 0}^{d\times d}$ and captures the geometry of $\M$ in $\Z$. This is known as the \emph{pullback metric} as it pulls the Euclidean metric from $\X$ into $\Z$. As the Jacobian spans the $d$-dimensional tangent space at the point $\x = f(\z)$, we may interpret $\b{M}(\z)$ as an inner product $\inner{\vec{u}}{\vec{v}}_{\b{M}} = \vec{u}\T\b{M}(\z)\vec{v}$ over this tangent space, given us all the ingredients to define \emph{Riemannian manifolds}:

\begin{figure}
\centering
    \includegraphics[width=0.48\linewidth]{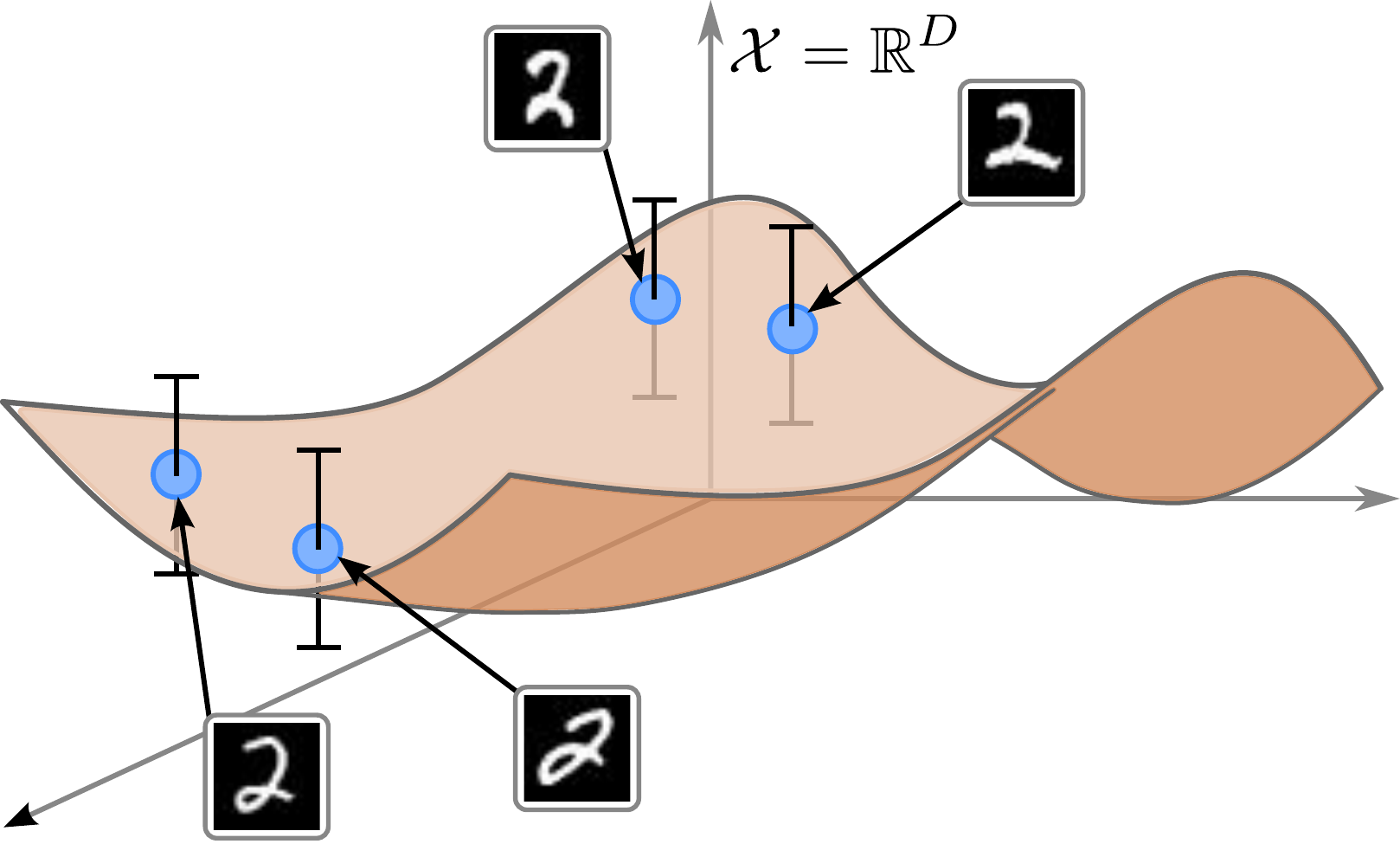}
    ~
    \includegraphics[width=0.48\linewidth]{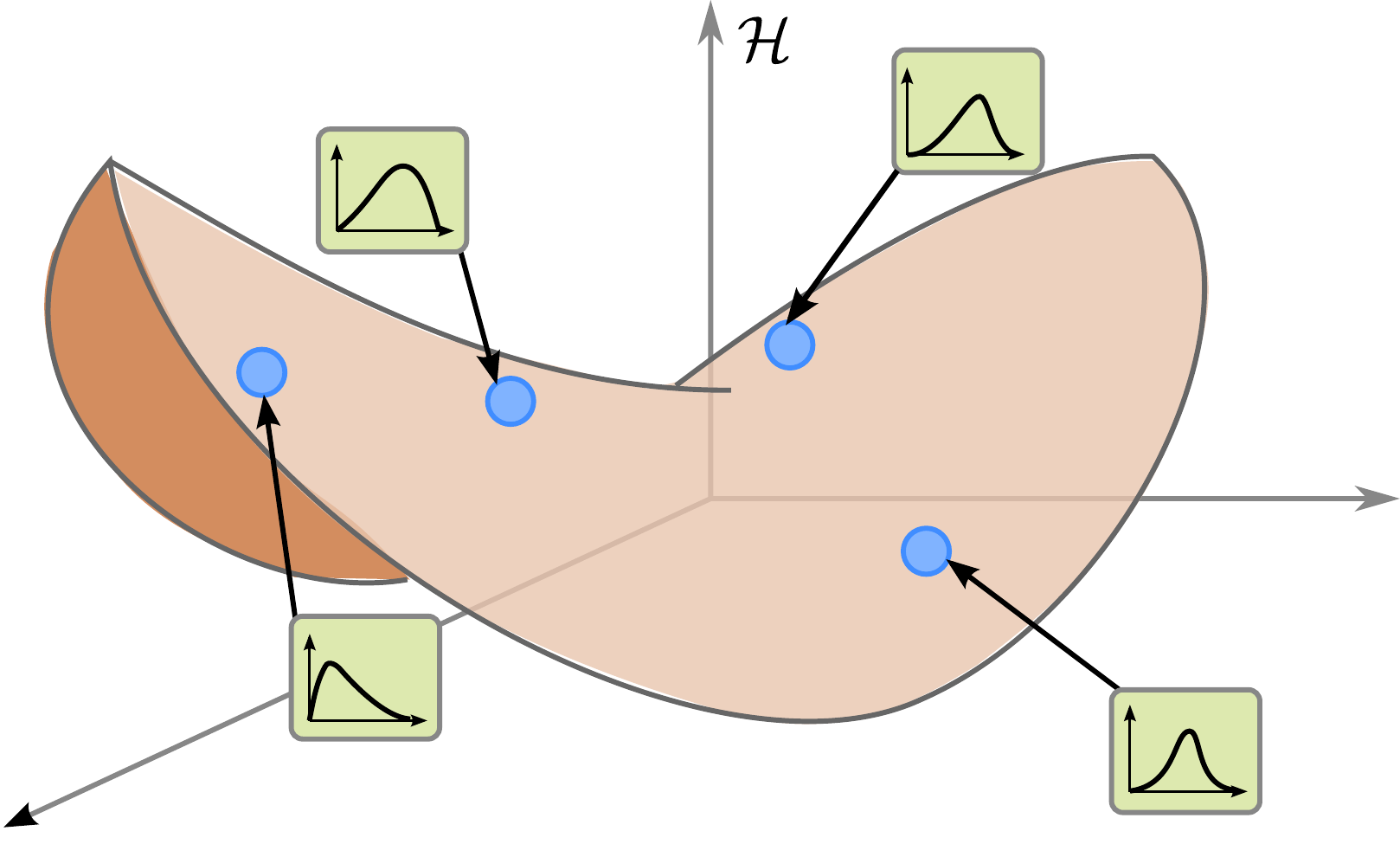}
    \caption{Traditionally (left), we view the learned manifold as a stochastic manifold in the observation space. We propose (right) to view the learned manifold as a deterministic manifold embedded in the space of decoder distributions, which is equipped with a Fisher-Rao metric based on information geometry.}
    \label{fig:teaser_img}
\end{figure}

  \begin{definition}
    A Riemannian manifold is a smooth manifold $\M$ together with a Riemannian metric $\b{M}({\z})$, which is a positive definite matrix that changes smoothly throughout space and defines an inner product on the tangent space $\mathcal{T}_{\z}\M$. 
  \end{definition}
  
We see that the decoder naturally spans a Riemannian manifold and the latent space $\Z$ can be considered as the \emph{intrinsic coordinates}. Technically, we can consider any Euclidean space as the intrinsic coordinates of an abstract $\M$ using a suitable metric $\b{M}(\z)$, which is implicitly induced by an abstract $f$. Since the Riemannian length of a latent curve \eqref{eq:curve_length}, by construction, is invariant to reparametrizations, it is natural to extend this view with a notion of \emph{distance}. We say that the distance between two points $\z_0, \z_1 \in \Z$ is simply the length of the shortest connecting path, $\mathrm{dist}(\z_0, \z_1) = \min_{\gamma} \mathrm{L}(\gamma)$. Calculating distances implies finding the shortest path. One can show \citep{gallot:2004} that length minimizing curves also have minimal \emph{energy}:
  \begin{align}
    \mathrm{E}(\gamma) 
      \!&= \!\int_0^1 \! \| \dot{f}(\gamma(t)) \|^2 \dif t \!= \!\int_0^1 \!\dot{\gamma}(t)\T \b{M}({\gamma(t)}) \dot{\gamma}(t) \dif t,
      \label{eq:curve_energy}
  \end{align}
  which is a locally convex functional. Shortest paths can then be found by direct energy minimization \citep{yang:arxiv:2018} or by solving the associated system of ordinary differential equations (ODEs) \citep{hennig:aistats:2014, arvanitidis:aistats:2019} (see supplementary materials for additional details).

\subsection{Stochastic decoders}

\begin{figure}
\centering
    \begin{overpic}[width=0.45\linewidth]{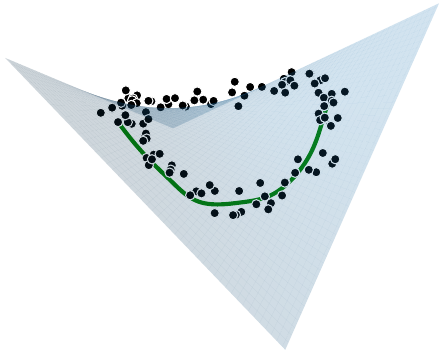}
    \put(0, 80){$\M\subset\X$}
    \put(22,22){\color{black}\vector(1,1){20}}
    \put(10, 10){$\mu(\gamma(t))$}
    \end{overpic}
    \qquad
    \begin{overpic}[width=0.35\linewidth]{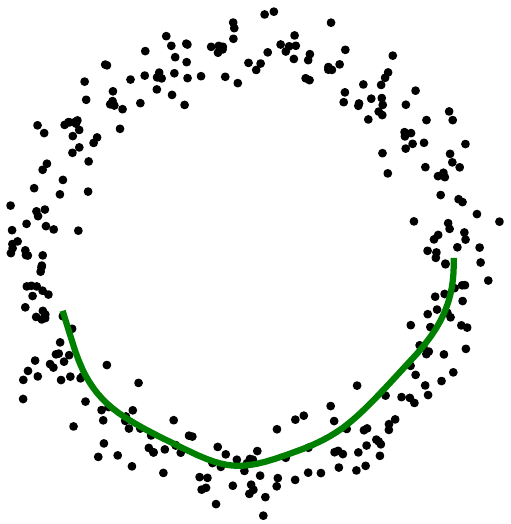}
    \put(0, 80){$\Z$}
    \put(45,50){\color{black}\vector(1,-1){20}}
    \put(25, 50){$\gamma(t)$}
    \end{overpic}
    \caption{A conceptual example of a Riemannian manifold $\M=\mu(\Z)$ lying in $\X$ and the corresponding latent space $\Z$, together with an associated shortest paths.}
    \label{fig:riem_geom_example}
\end{figure}
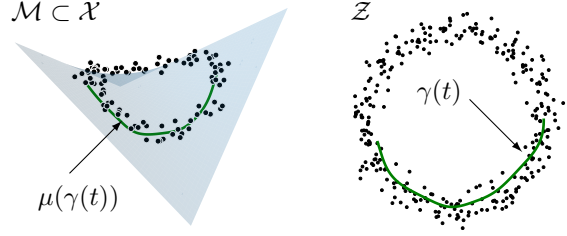

As previously discussed, deterministic decoders directly induce a Riemannian geometry in the latent space. However, most models of interest are stochastic and there is significant evidence that this stochasticity is important to faithfully capture the intrinsic structure of data \citep{hauberg:only:2018}. When the decoder is a smooth stochastic process, e.g.\ as in the Gaussian Process Latent Variable Model (GP-LVM) \citep{lawrence:2005:jmlr}, \citet{tosi:uai:2014} laid the foundations for modeling a stochastic geometry. Most contemporary models, such as VAEs, assume independent noise, making this theory inapplicable. \citet{arvanitidis:iclr:2018} proposed an extension of this stochastic geometry to VAEs with Gaussian decoders, which take the form
\begin{align}
    f(\z)
      &= \mu(\z) + \sigma(\z) \odot \vec{\epsilon} \nonumber\\
      &= \begin{bmatrix}
           \I_{D} & \mathrm{diag}(\vec{\epsilon})
         \end{bmatrix}
         \begin{bmatrix}
           \mu(\z) \\ \sigma(\z)
         \end{bmatrix}
      = \mat{P}_\varepsilon~ h(\z),
\end{align}
where $\vec{\epsilon} \sim \mathcal{N}(\vec{0}, \I_{D})$.
Here we have written the Gaussian decoder in its reparametrized form. This can be viewed as a random projection of a deterministic manifold spanned by $h$ with projection matrix $\b{P}_\epsilon$ \citep{eklund:arxiv:2019}, which can easily be given a geometry. The associated Riemannian metric,
\begin{align}
\label{eq:vae_metric}
    \b{M}({\b{z}}) = \b{J}_\mu(\z)\T\b{J}_\mu(\z) + \b{J}_\sigma(\z)\T\b{J}_\sigma(\z),
\end{align}
gives shortest paths that follow the data as distances grow with the model uncertainty \citep{arvanitidis:iclr:2018, hauberg:only:2018}. An example of a shortest path $\gamma(t)\in\Z$ computed under this metric is shown in Fig.~\ref{fig:riem_geom_example} and the respective curve on the corresponding expected manifold $\mu(\gamma(t))\in\M\subset\X$. 

Previous work has, thus, focused on \emph{pulling back} the Euclidean metric from the observation space to the latent space using the reparametrization of the Gaussian decoder. This is, however, intrinsically linked with the simple reparametrization of the Gaussian, and this strategy can only extend to location-scale distributions. We propose an alternative, principled way of dealing with stochasticity by changing the focus from the observation space $\X$ to the parameter space $\HH$ associated to the distribution of the decoder, leveraging the metrics defined in classical information geometry.

\section{Information geometric latent metric}
\label{sec:information_geometry}

So far we have seen how we can endow the latent space $\Z$ with meaningful distances only when our stochastic decoders are reparameterizable and their codomain is the observation space $\X$. Ideally, we would like a more general framework of computing shortest path distances for a more general class of distributions.

We first note that the codomain of a VAE decoder is the parameter space $\HH$ of a probability density function. In particular, depending on the type of data we specify a likelihood $p(\x|\eta)$ with parameters $\eta\in\HH$, which we can rewrite as $p(\x|\z)$ using the mapping $h:\Z\rightarrow\HH$.

With this in mind, we can ask what is a natural distance in the latent space $\Z$ between two infinitesimally near points $\z_1$ and $\z_2 = \z_1\!+\!\epsilon$ when measured in $\HH$. Since our latent codes map to distributions we can define the (infinitesimal) distance through the KL-divergence:
\begin{equation}
    \mathrm{dist}^2(\z_1, \z_2) = \KL(p(\x|\z_1),~p(\x|\z_2)).
    \label{eq:kl_dist}
\end{equation}

So we can define the length of a curve $\gamma\!:\![0, 1]\!\rightarrow\!\Z$ as
\begin{align}
  \mathrm{L}(\gamma)
    &= \lim_{N \rightarrow \infty} \sum_{n=1}^{N-1} {\KL(p(\x|\gamma(t_n)), p(\x|\gamma(t_{n+1})))}^{\frac{1}{2}},
\end{align}
and distances could be defined as before.
This would satisfy our desiderata of a deterministic notion of similarity in the latent space that is applicable to wide range of decoder distributions.

This construction may seem arbitrary, but in reality it carries deeper geometric meaning. \emph{Information geometry} \citep{nielsen:2020} considers families of probabilistic densities $p(\x|\eta)$ as represented by their parameters $\eta \in \HH$, such that $\HH$ is constructed as a statistical manifold equipped with the Fisher-Rao metric, which infinitesimally coincides with the KL divergence in \eqref{eq:kl_dist}. This is known to be a Riemannian metric over $\HH$ that takes the following form:

\begin{equation}
\label{eq:fisher_rao_metric}
    \b{I}_{\HH}(\eta)\!=\!\int_\X \![\nabla_{\eta} \log p(\b{x}|\eta)\nabla_{\eta} \log p(\b{x}|\eta)\T] p(\b{x}|\eta)\dif \b{x}.
\end{equation}
When the parameter space $\HH$ is equipped with this metric, we call it a \emph{statistical manifold}.

\begin{definition}
  A statistical manifold consists of the parameter space $\HH$ of a probability density function $p(\b{x}|\eta)$ equipped with the Fisher-Rao information matrix $\b{I}_{\HH}(\eta)$ as a Riemannian metric.
\end{definition}

Note that the geometry induced by the Fisher-Rao metric is predefined and can be seen as a modeling decision, since it is related to the chosen likelihood and does not change with data.

As previously mentioned, a known result in Information Geometry is that the Fisher-Rao metric coincides with the KL-divergence locally  \citep{nielsen:2020, amari:2016}:
\begin{proposition}
  The Fisher-Rao metric is the second order approximation of the KL-divergence between perturbed distributions: 
  \begin{equation}
  \label{eq:KL_approx_inner_product}
    \KL(p(\b{x}|\eta),p(\b{x}|\eta+\delta\eta)) = \frac{1}{2} \delta\eta\T \b{I}_{\HH}(\eta) \delta\eta + o(\delta\eta^2).
  \end{equation}
\end{proposition}
The central idea put forward in this paper is to consider the decoder as a map $h: \Z \rightarrow \HH$ instead of $f:\Z\rightarrow\X$, and let $\HH$ be equipped with the appropriate Fisher-Rao metric. The VAE can then be interpreted as spanning a manifold $h(\Z)$ in $\HH$ and the latent space $\Z$ can be endowed with the corresponding metric. We detail this approach in the sequel.

\subsection{The Riemannian pull-back metric}
Our construction implies that the length of a latent curve $\gamma:[0,1]\rightarrow \Z$ when mapped through $h$ can be measured in the parameter space $\HH$ using the Fisher-Rao metric therein as
\begin{align}
    \mathrm{L}(\gamma)
      &= \int_0^1 \sqrt{\partial_t h(\gamma(t))\T \b{I}_{\HH}(h(\gamma(t))) \partial_t h(\gamma(t))} \dif t,
\end{align}
  %

with $\b{M}$ the pullback metric:
\begin{proposition}
    \label{prop:fisher_pullback}
    Let $h:\Z\rightarrow\HH$ be an immersion that parametrizes the likelihood. Then, the latent space $\mathcal{Z}$ is equipped with the Riemannian pull-back metric $\b{M}(\b{z}) = \b{J}_{h}\T(\b{z}) \b{I}_{\HH}(h(\b{z})) \b{J}_{h}(\b{z})$.
\end{proposition}
\begin{proof}
See appendix, Prop. \ref{prop:riemannian-pullback}.
\end{proof}
  
Note that instead of considering the parameters $\eta\in\HH$ of the probabilistic density function $p(\x|\eta)$ that approximates the data, we can consider the latent variable $\b{z}$ as the actual parameters of the model. This view is equivalent to the one explained above, and the corresponding pull-back metric is directly the Fisher-Rao metric endowed in the latent space $\Z$:
\begin{proposition}
    The pullback metric $\b{M}(\b{z})$ is identical to the Fisher-Rao metric obtained over the parameter space $\Z$ as $\b{M}(\b{z}) = \int_{\X} \left[\nabla_\b{z}\log p(\b{x}|\b{z}) \nabla_\b{z} \log p(\b{x}|\b{z})\T\right] p(\b{x}|\b{z}) \dif\b{x}. $
\end{proposition}
\begin{proof}
See appendix, Prop. \ref{prop:pullback-FR-equal}.
\end{proof}

Therefore, pulling back the Fisher-Rao metric from $\HH$ into $\Z$ enables us to compute length minimizing curves which are indentifiable (see Sec. \ref{sec:riemannian_geometry}). The advantange of this approach is that it applies to any type of decoders and data, as the actual distance is measured over the manifold spanned by $h$ in the parameter space $\HH$. So shortest paths between probability distributions move optimally on this manifold while taking the geometry of $\HH$ into account through the Fisher-Rao metric.

Computing shortest paths directly in $\HH$ need not result in a sensible sequence of probability density functions $p(\b{x}|\eta)$. To ensure that the shortest paths computed under our metric stay within the support of the data, we carefully design our decoder $h$ to extrapolate to uncertain distributions outside the support of the data (see supplements for additional details).

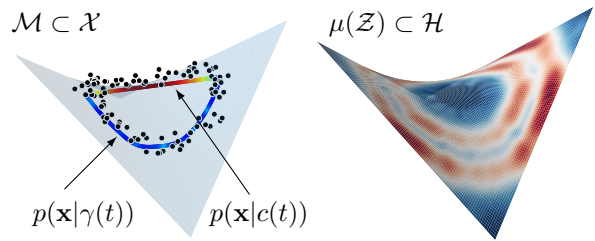
\begin{figure}
\centering
    \begin{overpic}[width=0.45\linewidth]{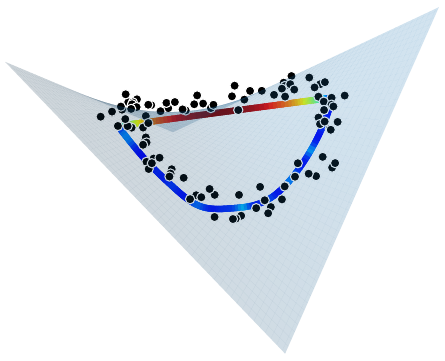}
    \put(0, 80){$\M\subset\X$}
    \put(20,20){\color{black}\vector(1,1){20}}
    \put(90,18){\color{black}\vector(-0.7,1){28}}
    \put(8, 8){$p(\x|\gamma(t))$}
    \put(75, 8){$p(\x|c(t))$}
    \end{overpic}
    ~
    \begin{overpic}[width=0.45\linewidth]{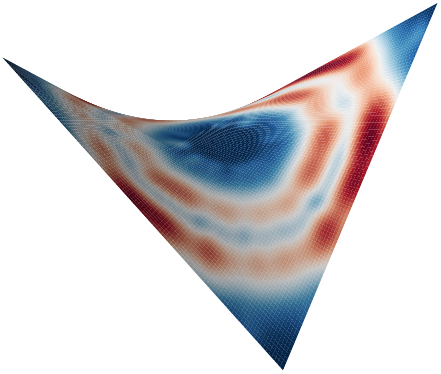}
    \put(5, 80){$\mu(\Z)\subset\HH$}
    \end{overpic}

    \caption{\emph{Left}: The optimal $\gamma(t)$ under $\b{M}(\z)$ results to distributions that respect the structure of data, while the curve $c(t)$ with minimal length in $\HH$ does not as it leaves $\M$. Red and green signal high and low variance respectively. \emph{Right}: A part of the spanned manifold $h(\Z) = [\mu(\Z), \sigma(\Z)] \in\HH$ colored by $|\mathbf{M}(\z)|$. Note that we design $\sigma(\z)$ to increase far from data, which ensures that $\gamma(t)$ stays within their support.}
    \label{fig:inf_geom_compare_ours_vs_fisher}
\end{figure} 

In Fig.~\ref{fig:inf_geom_compare_ours_vs_fisher} we compare a shortest path  $\gamma\!:\![0,1]\!\rightarrow\!\Z$ under the proposed metric $\b{M}(\z)$ against a curve $c\!:\![0,1]\!\rightarrow\!\HH$ with minimal length. We consider a Gaussian likelihood with isotropic covariance. We show the resulting sequence of means for both interpolants color-coded by the corresponding variances. As expected $c(t)$ does not take into account the given data, but only respects the geometry of $\HH$ implied by the likelihood.

\subsection{Efficient shortest path computation}
\label{sec:geodesic_computation}

  An essential task in computational geometry is to compute shortest paths.
  This can be achieved by minimizing curve energy \eqref{eq:curve_energy} or solving the corresponding system of ODEs (see supplementary material). The latter, however, requires inordinate computational resources, since the evaluation of the system relies on the Jacobian of the decoder and its derivatives. 
  
  Bearing in mind that the metric is an approximation of the KL divergence between perturbations \eqref{eq:KL_approx_inner_product}, the energy is directly expressed as a sum of KL divergence terms along a discretized curve $\gamma$:
  \begin{equation}
    \label{eq:our_energy_with_kl}
        \mathrm{E}(\gamma) \propto \lim_{N \rightarrow \infty}\sum_{n=1}^{N-1} \KL(p(\x|\gamma(t_n)), p(\x|\gamma(t_{n+1}))).
  \end{equation}
  The proof can be found in the appendix, Prop. \ref{prop:FRenergydistance}. A simple algorithm for computing shortest paths is to minimize \eqref{eq:our_energy_with_kl} with respect to the parameters of the curve $\gamma$. Here we represent $\gamma$ as a cubic spline with fixed end-points. Then standard free-form optimization can be applied to minimize this energy.

\subsection{Example: categorical decoders}
\label{sec:example_categorical}

The motivation for our approach is that, while several options for decoders exist in VAEs depending on the type of the given data, we could only capture and use the learned geometry in a principled way with Gaussian decoders. Our proposed methodology is more general.

For a constructive example, assume that $\x$ is a categorical variable. We can select a generalized Bernoulli likelihood $p(\x|\z)$, such that $h(\z) = (\eta_1, \cdots, \eta_D)$ where each $\eta_i$ represents the probability of $x_i$ being 1. Thus, the parameters $\eta$ lie on the unit simplex $\HH$, and the distance under the corresponding Fisher-Rao metric between points on the simplex coincides with the spherical distance between the points $\sqrt{\eta}$ on the unit sphere, 
\begin{equation}
    \mathrm{dist}(\eta,\eta') = \mathrm{arccos}\left(\sqrt{\eta}\T\sqrt{\eta'}\right).
\end{equation}
We derive in detail this previously known result in the supplementary materials. 

Given a curve $\gamma:[0,1]\rightarrow\Z$ we can approximate the energy by using the small angle approximation $\cos\theta \approx 1 - \sfrac{\theta^2}{2} \Leftrightarrow \theta^2 \approx 2 - 2\cos\theta$ to give
\begin{equation}
    \mathrm{E}(\gamma) = \sum_{n=1}^{N-1}\left(2 - 2\sqrt{h(\gamma(t_n))}\T\sqrt{h(\gamma(t_{n+1}))}\right),
\end{equation}
for sufficiently fine discretization with $t_n = \sfrac{n}{N}$ and $t_{n+1} = \sfrac{n+1}{N}$. This gives a particular simple expression for the energy, which we can minimize in order to compute the shortest path.

\subsection{Black-box random geometry}\label{sec:blackbox}
  In general, we can derive suitable expressions for computing metrics and energies for families of decoders, doing so is tedious, error-prone and time-consuming. This limits the practical use of the developed theory.
  
  Drawing inspiration from \emph{black-box variational inference} \citep{ranganath:aistats:2014}, we propose a notion of \emph{black-box random geometry}. Assume that we have access to a differentiable KL divergence for our choice of decoder distribution. We can then apply the methodology presented in Sec.~\ref{sec:geodesic_computation} to compute shortest paths. 
  
  In practice, modern libraries such as PyTorch \citep{paszke:neurips:2019} have this functionality implemented for several distributions. When we do not have closed-form expression for the KL divergence, we can resort to Monte Carlo estimates thereof. More specifically, we can estimate the KL divergence by generating samples from the likelihood based on the re-parametrization trick, which allows us to get derivatives with automatic differentiation. 

  Interestingly, apart from finding the shortest path through the KL formulation, we can also approximate the actual metric tensor $\b{M}(\z)$. As we have discussed above, evaluating explicitly this metric is not a trivial task in many cases. One problem is that we need access to the Jacobian of the parametrization $h$, which is typically a deep neural network, so the computation is not always straightforward. Alternatively, one could use that the Fisher-Rao metric is the Hessian of the KL-divergence \eqref{eq:KL_approx_inner_product}, but such approaches fare poorly with current tools for automatic differentiation, where higher-order derivatives are often incompatible with batching. Furthermore, the Fisher-Rao metric itself may be intractable depending on the chosen likelihood $p(\x|\eta)$. Nevertheless, we show that the KL formulation \eqref{eq:KL_approx_inner_product} allows us to approximate the latent metric as:
    \begin{proposition}
    \label{prop:approximating_metric}
    We define perturbations vectors as $\delta e_{i} = \varepsilon \cdot \b{e_i}$, with $\varepsilon\in\mathbb{R}_{+}$ a small infinitesimal quantity, and $\b{e_i}$ a canonical basis vector in $\mathbb{R}^d$. For better clarity, we rename $\mathrm{KL}(p(\x|\z), p(\x|\z+\delta \z)) = \mathrm{KL}_{\z}(\delta \z)$ and we note $\b{M}_{ij} = \b{M}_{ji}$ the components of $\b{M}(\z)$.
    We can then approximate by a system of equations the diagonal and non-diagonal elements of the metric:
        \begin{align*}
            \b{M}_{ii} &\approx 2 \ \mathrm{KL}_{\z}(\delta \b{e_i})/ \varepsilon^2 \ \\
            \b{M}_{ji} &\approx  \left(\mathrm{KL}_{\z}(\delta \b{e_i}\!+\! \delta \b{e_j})- \mathrm{KL}_{\z}(\delta \b{e_i}) -\mathrm{KL}_{\z}(\delta \b{e_j}) \right)/ \varepsilon^2. 
        \end{align*}
    \end{proposition}
  See Prop. \ref{appendix:prop:metric_approximation} in the appendix for a proof. Note that this formulation only requires $h$ to be a smooth immersion. This is particularly useful, as the metric is used for other purposes on a Riemannian manifold and not exclusively for computing shortest paths. For example, relying on $\b{M}(\z)$ we can compute the exponential map by solving the corresponding ODE system as an initial value problem. Assuming a fully differentiable KL divergence, then the approximated metric is also differentiable. This is all that is required for practical usage of differential geometry, and thus, we have a reasonable notion of \emph{black-box random geometry}.

\section{Experiments}
\label{sec:experiments}

 \begin{figure*}[t!]
      \resizebox{\textwidth}{!}{
      \begin{tabular}{cccc}
        \includegraphics[height=3.5cm]{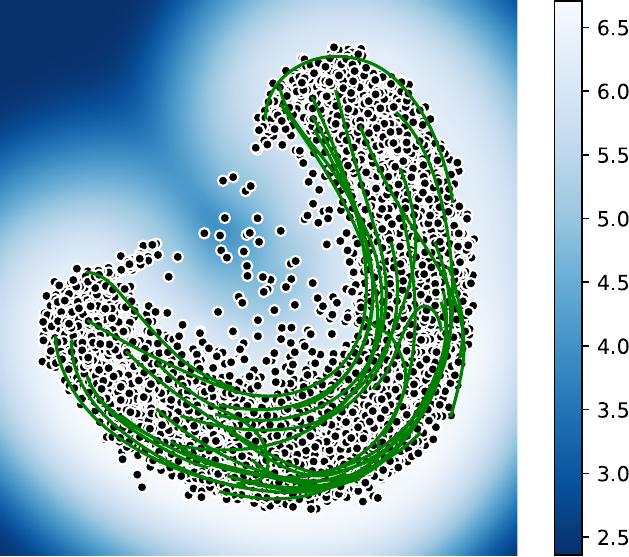} &
        \includegraphics[height=3.5cm]{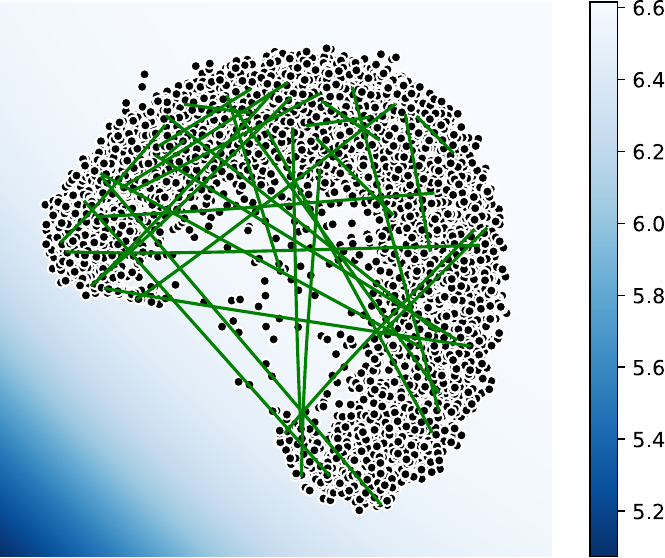} &
        \includegraphics[height=3.5cm]{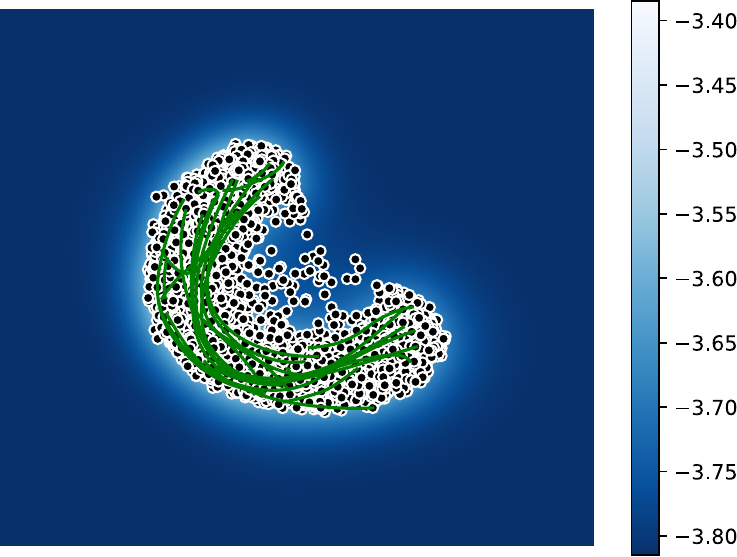} &
        \includegraphics[height=3.5cm]{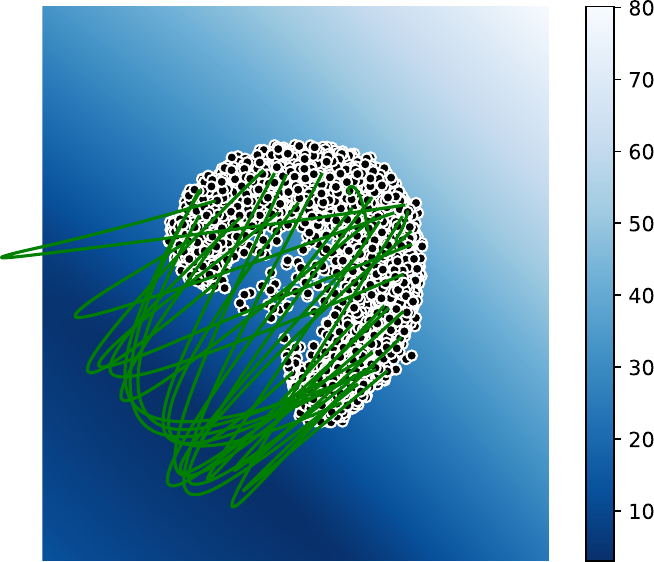} \\
      \end{tabular} }
      \caption{Pulling back the Euclidean and Fisher-Rao metrics with Gaussian decoders. Left to right: Euclidean pull-back with regularized uncertainty, Euclidean pull-back with a NN to model uncertainty, Fisher-Rao pull-back with regularized uncertainty, Fisher-Rao pull-back with a NN to model uncertainty.}
      \label{fig:gaussian_decoders_UQ}
  \end{figure*}

\begin{figure*}[t!]
    \centering
    \begin{subfigure}{0.19\textwidth}
	    \includegraphics[width=1\linewidth]{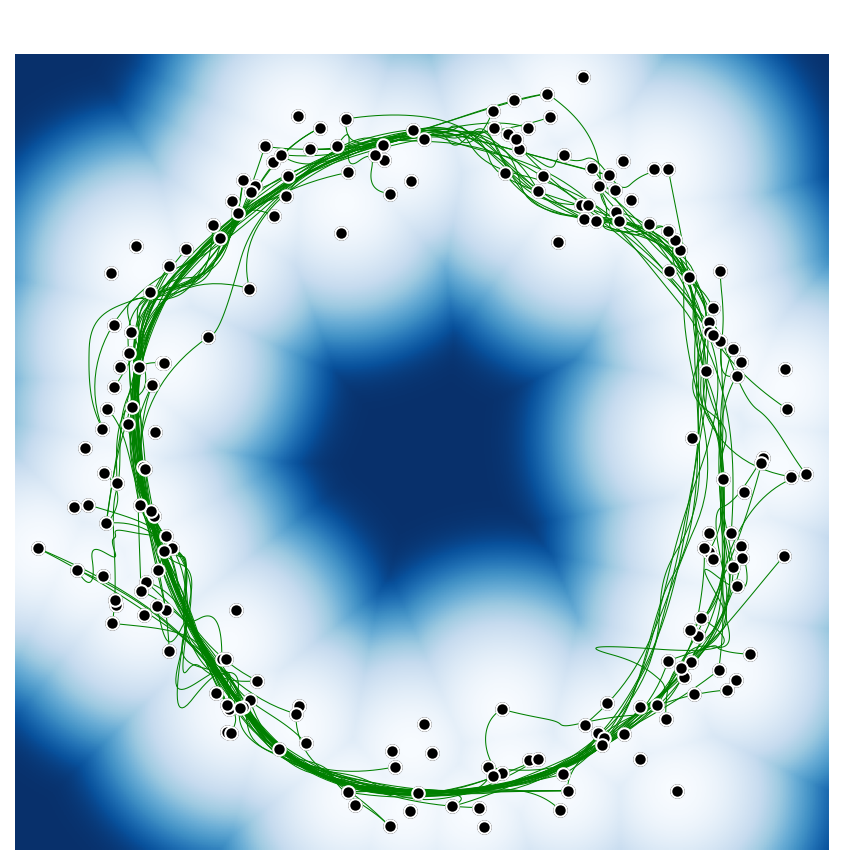}
    \end{subfigure}
    \begin{subfigure}{0.19\textwidth}
	    \includegraphics[width=1\linewidth]{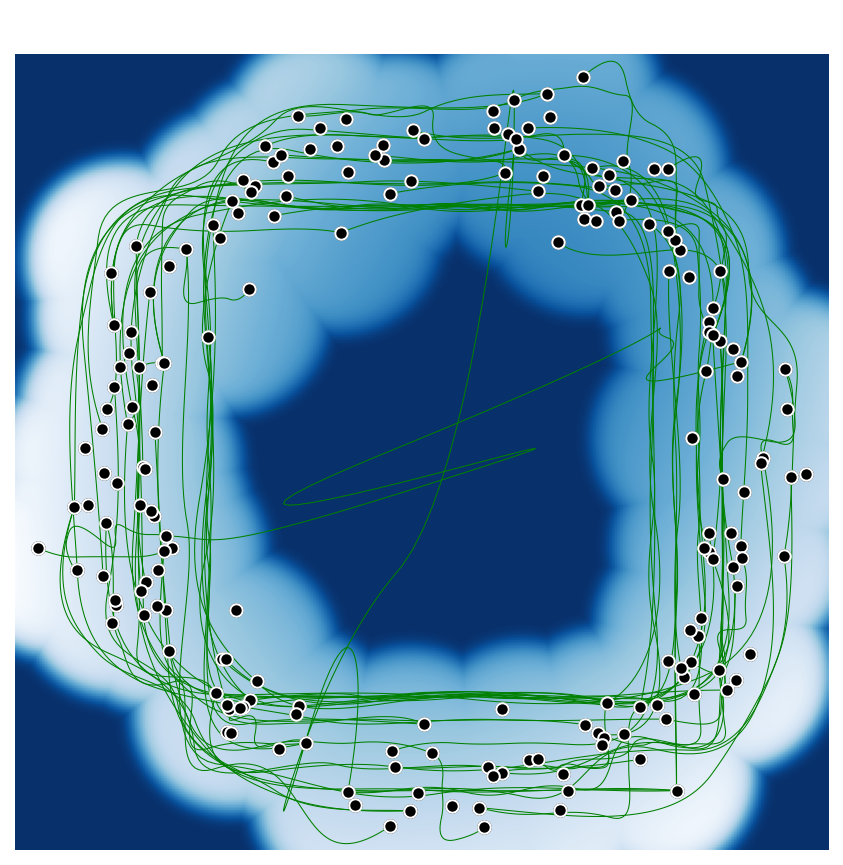}
    \end{subfigure}
    \begin{subfigure}{0.19\textwidth}
	    \includegraphics[width=1\linewidth]{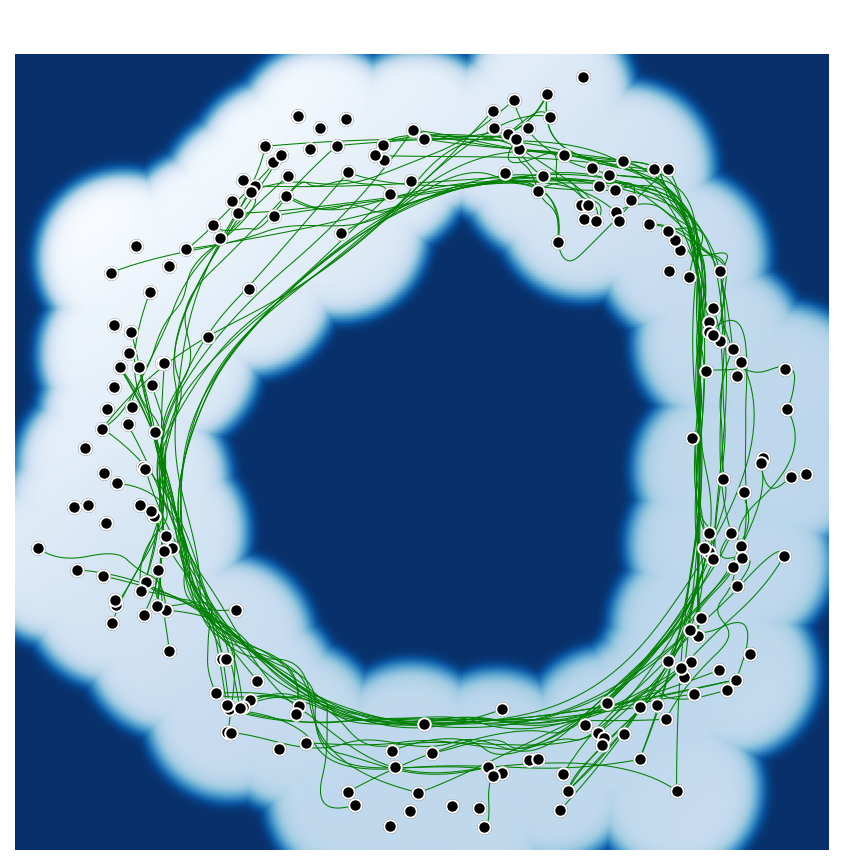}
    \end{subfigure}
    \begin{subfigure}{0.19\textwidth}
	    \includegraphics[width=1\linewidth]{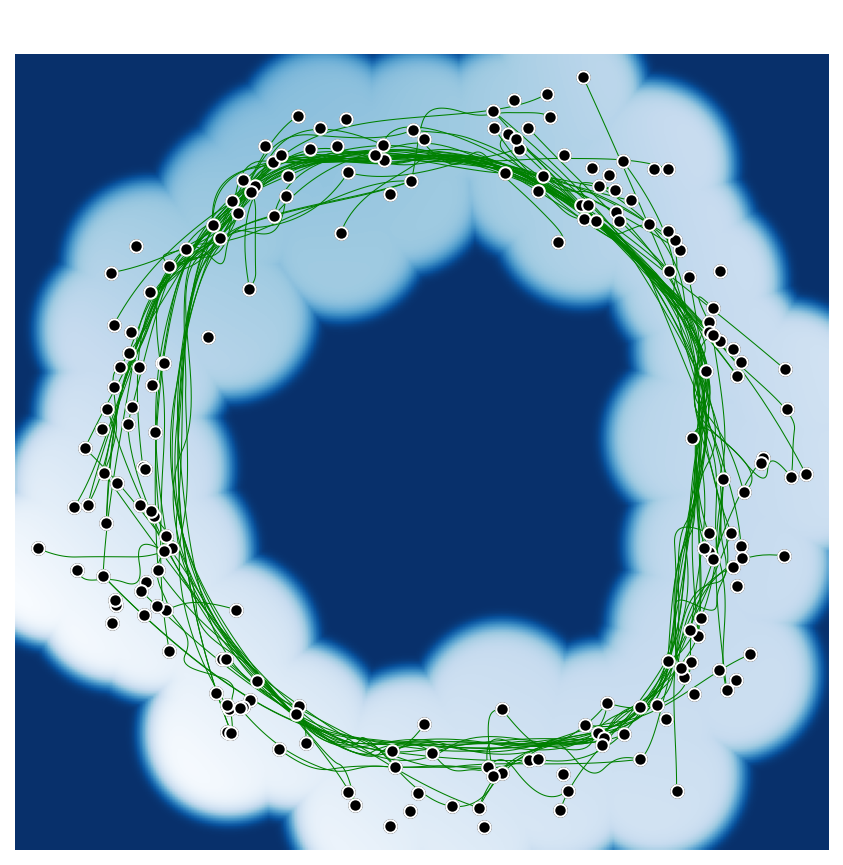}
    \end{subfigure}
    \begin{subfigure}{0.19\textwidth}
	    \includegraphics[width=1\linewidth]{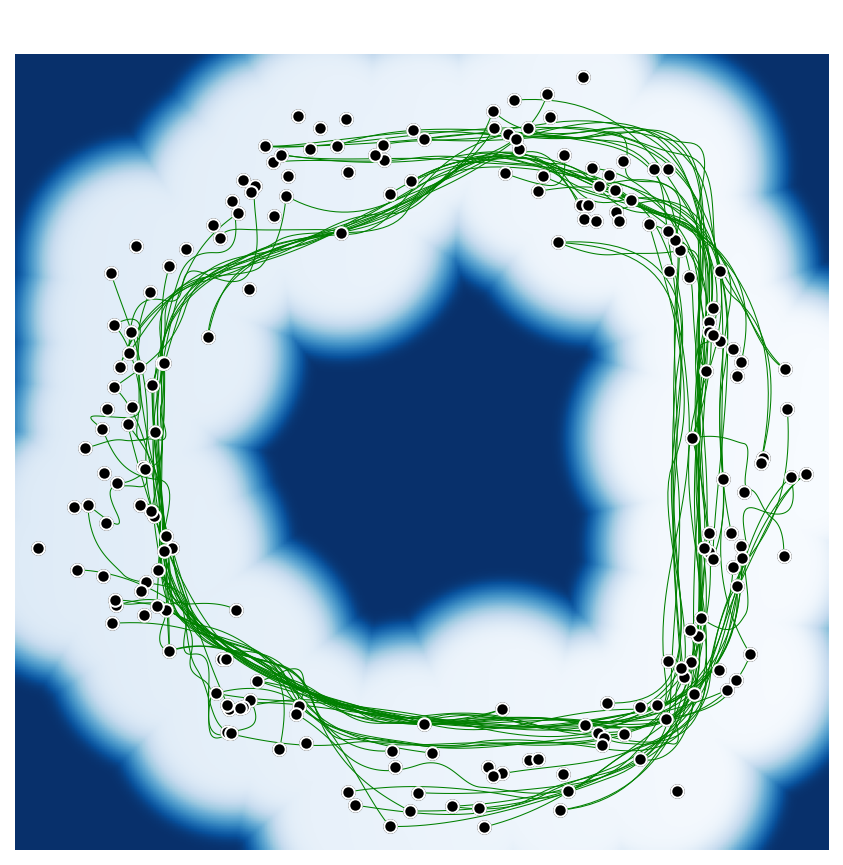}
    \end{subfigure}
    \caption{Pulling back the metric from different parameter spaces. From left to right: Normal, Bernoulli, Beta, Dirichlet and Exponential. White areas represent low entropy of the decoded distribution, while blue areas represent higher entropy. Notice that the Bernoulli latent space is darker blue (i.e.\@ more entropic) because distributions with parameters around $\sfrac{1}{2}$ are near uniform.}
    \label{fig:latent_spaces_toy_experiment}
\end{figure*}

\subsection{Pulling back Euclidean and Fisher-Rao metric with Gaussian decoders}
\label{sec:compare_metrics_with_gaussians_generators_uq}

We start our experiments by comparing our proposed way of inducing geometry in latent spaces with the existing theory: pulling back the Euclidean metric using a stochastic Gaussian decoder (see \eqref{eq:vae_metric}). We also include in this comparison the effect of regularizing the uncertainty quantification in the learned geometries. In this regularization, we use transition networks \citep{detlefsen:2019:reliable} to ensure high uncertainty outside the support of the data (see Sec.~\ref{appendix:sec:UQ} in the supplementary material).

In this experiment, we train four VAEs on a subset of the MNIST dataset composed of only the digits with label 1. Two of these VAEs implement a standard Gaussian decoder, and we induce a metric in the latent space by pulling the Euclidean metric back using the Jacobian of the decoder. In the other two, we consider the output of the decoder as lying in a statistical manifold and approximate the pullback of the Fisher-Rao metric by using the KL divergence locally. In each of these two sets, one of the decoders implements the uncertainty regularization described above.

Fig.~\ref{fig:gaussian_decoders_UQ} shows the latent spaces of these four decoders, illuminated by the volume measure. In each of this latent spaces, we analyze the geometry induced by the respective pullbacks by computing and plotting several shortest paths. This figure illustrates two key findings: (1) Our approach is on par with the existing literature in learning geometric structure, which can be seen by comparing the first and third latent spaces (Euclidean vs. Fisher Rao, respectively), and (2) Performing uncertainty regularization plays an instrumental role on learning a sensible geometric structure, which can be seen when comparing the first and second latent spaces (both coming from the Euclidean pullback, with and without regularization respectively), and similarly for the third and fourth.

\subsection{The Fisher-Rao pullback metric for various distributions with toy data}
\label{sec:toy_latent_space_experiment}

For our second experiment, we induced a geometry on a known latent space (given by noisy circular data in $\mathcal{Z}_{\text{toy}} = \mathbb{R}^2$) by \textit{pulling back} the Fisher-Rao metric from the parameter space of different distributions, showcasing the potential for computing shortest paths efficiently, even in non-Gaussian settings. 
The statistical manifolds from which we pull the metric are associated with multivariate versions of the Normal, Bernoulli, Beta, Dirichlet and Exponential distributions. For this approximation to follow the support of the data we need to ensure that our mapping $\mathcal{Z}_{\text{toy}}\to\mathcal{H}$ extrapolates to high uncertainty outside our training codes (see Fig. \ref{fig:gaussian_decoders_UQ}). To do so, we perform uncertainty regularization for each one of the decoded distributions (see supplementary materials for implementation details).

\begin{figure*}
    \centering
    \includegraphics[height=3cm]{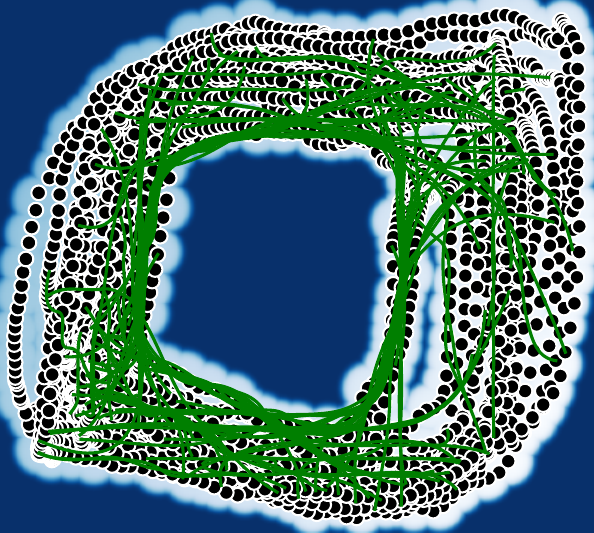}
    ~
    \includegraphics[height=3cm]{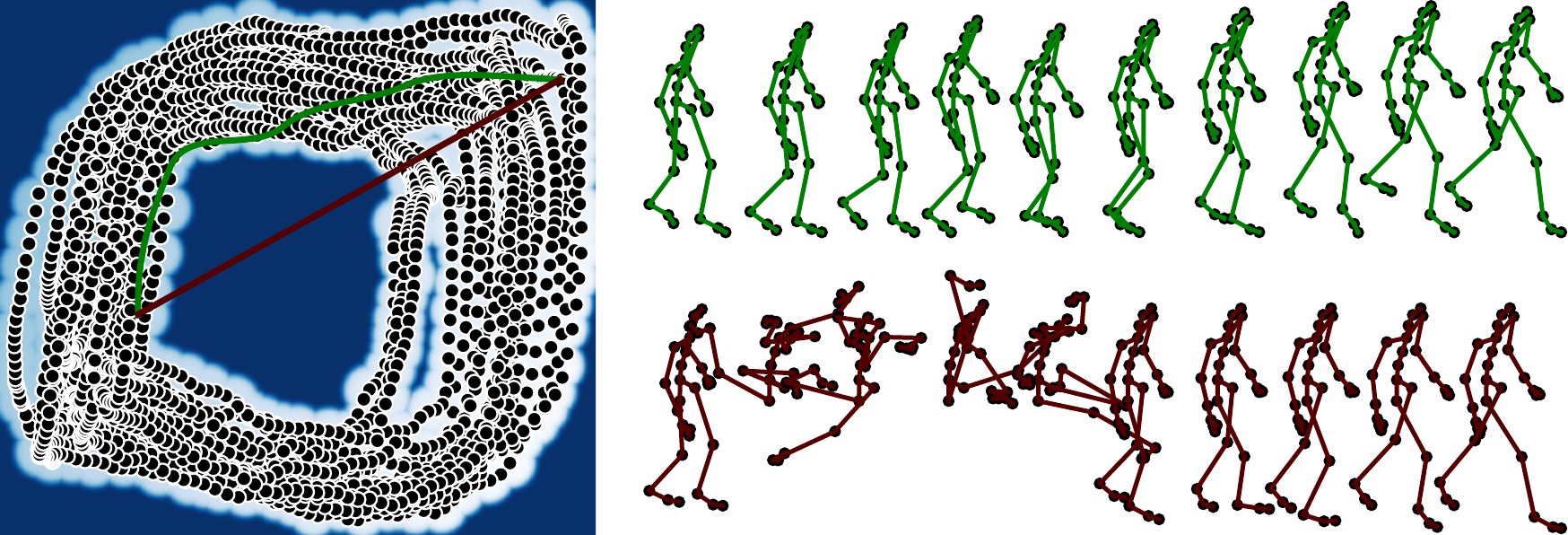}
    \caption{\textit{Left:} Geodesics in the latent space of a von Mises-Fisher decoder. \textit{Middle}: Shortest path (green) vs. linear (red). \textit{Right}: decoding the shortest path (green) vs. the linear interpolation (red) as poses (i.e. the product of von Mises-Fisher distributions). Our path follows the trend of the data manifold, while the linear path traverses regions with no data support.}
    \label{fig:mocap_interp}
\end{figure*}

In Fig.~\ref{fig:latent_spaces_toy_experiment} we show the toy latent space alongside several shortest paths computed using the pullback of the Fisher-Rao metric from the statistical manifolds associated with the Gaussian, Bernoulli, Beta, Dirichlet and Exponential distributions. We parametrize the curves as cubic splines and minimize their energy using automatic differentiation (see Sec. \ref{sec:geodesic_computation}). These results show that the approximated pulled-back metric induces a meaningful geometry in this latent space, which recovers the true circular structure of the data. In the case of the Bernoulli distribution, we notice that some of the paths fail to converge. We hypothesize that our uncertainty regularization (which decodes to the uniform distribution outside the support) is not strong enough since Bernoulli distributions with parameters close to $1/2$ are already highly entropic.


\subsection{Motion capture data with products of von Mises-Fisher distributions}
\label{sec:motion-capture}

As a further demonstration of our black-box random geometry, we consider a model of human motion capture data. Here we observe a time series, where each time point represent a `skeleton' corresponding to a human pose. As only pose, and not shape, changes over time, individual limbs on the body only change position and orientation, but not length. Each limb is then a point on a sphere in $\mathbb{R}^3$ with radius given by the limb length. Following \citet{tournier2009motion} we view the skeleton representation space as a product of spheres. From this, we build a VAE where the decoder distribution is a product of von Mises-Fisher distributions. To ensure a sensible uncertainty estimates in the decoder, we enforce that the concentration parameter extrapolate to a small constant.

In this case, we do not have easily accessible Fisher-Rao metrics, so we lean on the KL formulation from Sec.~\ref{sec:blackbox}. Since, the KL does not have a closed-form expression for the von Mises-Fisher distribution, we resort to a Monte Carlo estimate thereof. This is realisable with off-the-shelf tools \citep{s-vae18}.


Fig.~\ref{fig:mocap_interp} shows the latent representation of a motion capture sequence of a person walking ({Seq.\ \texttt{69\_06} from \url{http://mocap.cs.cmu.edu/}}) with shortest paths superimposed. We see that our paths follow the trend of the data, and reflect the underlying periodic nature of the observed walking motion. We pick two random points in the latent space, and traverse both the shortest path and the straight line implied by a Euclidean interpretation of the latent space. As we traverse, we sample from the decoder distribution, thereby producing two new motion sequences, which appear in Fig.~\ref{fig:mocap_interp}. As can be seen, the straight line traverses uncharted territory of the latent space and end up creating an implausible motion. This is in contrast to the shortest path, that consistently generates meaningful poses.


\subsection{Numerical approximation of the Fisher-Rao pullback metric}
\label{sec:experiment_numerical_approx_metric}

\begin{figure}
\centering
    \includegraphics[height=3.5cm]{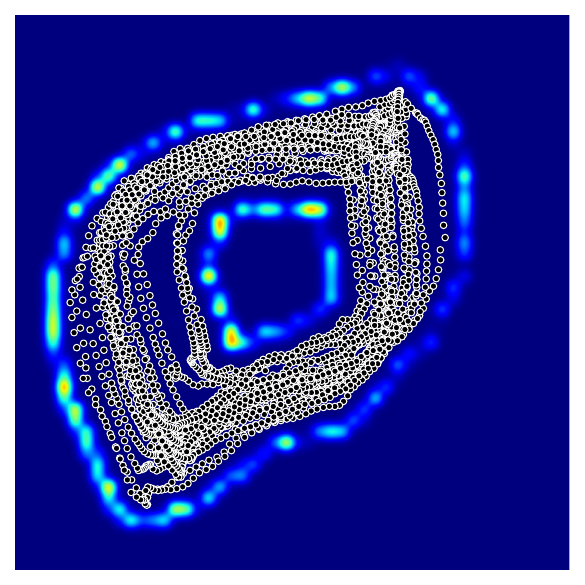}
    ~
    \includegraphics[width=3.5cm]{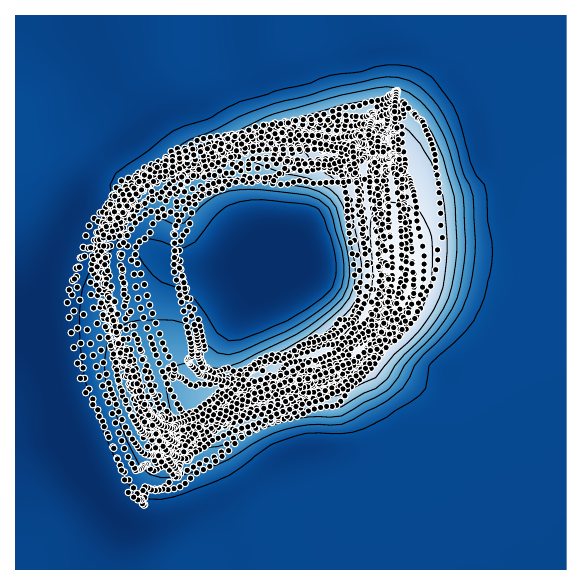}
    \caption{\emph{Left}: Average error of the approximated metric in the von Mises-Fisher latent space. Darker colors indicate lower error (less than $\varepsilon^2$), while higher values are clear. \emph{Right}: The LAND density well-adapts to the nonlinear structure of the latent representations due to the shortest paths behavior.}
    \label{fig:approx_error_and_land}
\end{figure}

Prop.~\ref{prop:approximating_metric} provide an approximation to the metric and we test its accuracy as per \eqref{eq:KL_approx_inner_product}. We discretize the latent space for the just-described von Mises-Fisher decoder and, for each $\z$ in this grid, we both approximate $\b{M}(\z)$ and compute the expected value of $\lVert\mathrm{KL}(p(\x|\z), p(\x|\z+\delta \z)) - \frac{1}{2}\delta\z^\top \b{M}(\z)\delta\z\rVert$ for several samples of $\delta\z$, uniformly distributed around the circle of radius $\varepsilon = 0.1$. Notice that we do not have a ground truth to compare against, and that this error will always be off by $o(\delta \z^2)$.
Fig.~\ref{fig:approx_error_and_land} shows the average error, where we can see that the approximate metric is well-estimated both within and outside the support of the data. The error, however, grows at the boundaries of the support, where the distribution is changing from a concentrated von Mises-Fisher to a uniform distribution.
It is worth mentioning that we observe some approximated metrics have negative determinant, showing that our numerical approximations are imprecise at the boundary. These results warrant further research on more stable ways of approximating pulled back metrics under our proposed approach.

\subsection{Statistical models on manifolds}
\label{sec:LAND_experiment}
We demonstrate the usefulness of the approximated metrics, by fitting a distribution to data in the latent space, which requires normalization according to the measure induced by the metric. In particular, we fit a locally adaptive normal distribution (LAND) \citep{arvanitidis:nips:2016}, which extends the Gaussian distribution to learned manifolds. The probability density function is $\rho(\b{z}) = C(\bs{\mu},\bs{\Sigma}) \cdot \exp\left(-0.5\cdot \mathrm{Log}_{\bs{\mu}}(\z)\T \bs{\Gamma}\mathrm{Log}_{\bs{\mu}}(\z) \right)$, where $\bs{\mu}\in\R^d$ is the mean, $\bs{\Gamma}\in\R_{\succ 0}^{d\times d}$ is the precision matrix and $C(\bs{\mu}, \bs{\Gamma})$ the normalization constant. The operator $\mathrm{Log}_{\bs{\mu}}(\z)$ returns the scaled initial velocity $\mathbf{v}=\dot{\gamma}(0)\in\R^d$ of the shortest connecting path with $\gamma(1)=\z$ and $||\mathbf{v}||=\mathrm{Length}(\gamma)$. In Fig.~\ref{fig:approx_error_and_land} we show the LAND density on the learned latent representations under the approximated Riemannian metric from Sec.~\ref{sec:experiment_numerical_approx_metric}. Since shortest paths follow the data, so does the density $\rho$. See supplementary material for details.

\subsection{Movie preferences via latent interpolants}
\label{sec:movie-ratings}

In addition, we explored the latent space of the movie-users rating  dataset MovieLens 25M (\url{https://grouplens.org/datasets/movielens/25m/}). In particular, we consider a Bernoulli VAE to model if a user has watched a movie among the 60 most popular in the dataset. Also, we considered only users who have seen less than 30 movies. The implementation and preprocessing details can be found in the supplementary material. Our VAE decodes to 60 Bernoulli parameters that are conditionally independent given the latent code $\z$, which state the likelihood that a given user has seen these movies. Latent codes in this space, then, can be seen as individual users with certain movie preferences.

\begin{figure}
    \centering
    \begin{subfigure}{0.45\linewidth}
        \includegraphics[height=3.5cm]{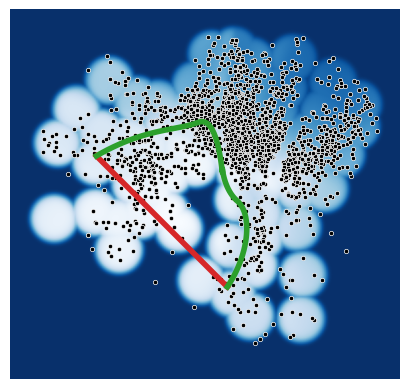}
    \end{subfigure}
    \begin{subfigure}{0.45\linewidth}
        \begin{overpic}[height=1.65cm]{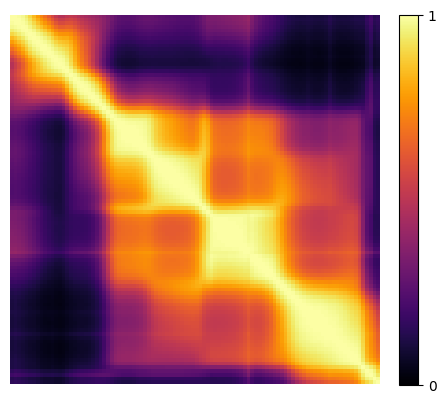}
        \put(12, -5){\tiny Our path}
        \end{overpic}
        \begin{overpic}[height=1.65cm]{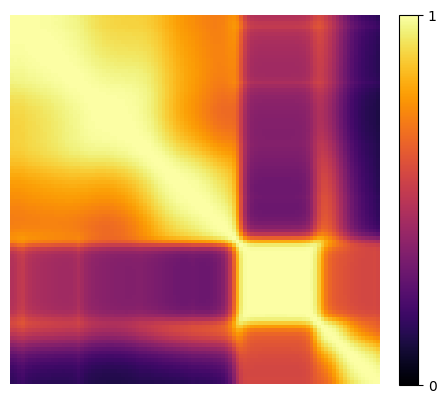}
        \put(7, -5){\tiny Linear path}
        \end{overpic}
        \vspace{2pt}
        \caption*{\small Cosine similarities}
    \end{subfigure}
    
    \caption{Our path (green) follows users with similar  preferences, as similarity is only locally high. Instead, the line (red) does not respect the learned structure resulting to users with no specific preferences.}
    \label{fig:movies_experiment}
\end{figure}

We then computed the shortest path between two points by considering the pulled-back Fisher-Rao (see Sec. \ref{sec:geodesic_computation}), and we compare against a straight line interpolation. We consider the cosine similarity of the decoded outputs. This cosine similarity measures whether two users (encoded as points in the latent space) have similar preferences according to our model. In Fig. \ref{fig:movies_experiment} we see that our path follows users with similar movie preferences locally, while the linear interpolation failed to capture a local notion of preference.

\section{Related work}
    The literature is rich on deterministic generative models such as autoencoders \citep{rumelhart:nature:1986} and generative adversarial networks \citep{goodfellow:neurips:2014},and a series of papers have investigated such deterministic decoders \citep{shao2018riemannian, chen2018metrics, laine2018feature}. However, our work is not applicable to this setting. As demonstrated in Sec.~\ref{sec:compare_metrics_with_gaussians_generators_uq} stochasticity is essential to shape the latent space according to the data manifold. \citet{hauberg:only:2018} argues model uncertainty plays a role much akin to topology in classic geometry, in that it, practically, allows us to deviate from the Euclidean topology of the latent space.
    
    Our constructions rely on information geometry and in particular Fisher-Rao metrics \citep{nielsen:2020}. While our work is within the spirit of information geometry, it does not represent typical usage of this theory. Information geometry has been widely used in the context of optimisation with \emph{natural gradients} \citep{martens2014new, martens2015optimizing}, Markov Chain Monte Carlo methods \citep{girolami:2011} and hypothesis testing \citep{nielsen:2020}. The key difference between natural gradients and our work is the space we wish to explore: in the case of the natural gradients, the shortest path is obtained on the space of the weights of the neural networks, while we aim to explore the latent space of a VAE. It can also be noted that Information geometry provides a rich family of alternative divergences over the here-applied KL-divergence. We did not investigate their usage in our context.
    
    To make use of the here-developed tools, we may lean on techniques for statistics on manifolds. These provide generalizations of a long list of classic statistical algorithms \citep{zhang2013probabilistic, hauberg:tpami:princurve, fletcher2011geodesic}. We refer the reader to \citet{pennec:jmiv:2006} for a gentle introduction to this line of research.

\section{Conclusion and discussion}
\label{sec:conclusion}
We have proposed a new approach for getting a well-defined and useful geometry in the latent space of generative models with stochastic decoders. The theory is easy to apply and readily generalize to a large family of decoder distributions. The latent geometry gives access to a series of operations on latent variables that are invariant to reparametrizations of the latent space, and therefore are not subject to a large class of identifiability issues. Such operational representations have already shown great value in applications ranging from biology \citep{detlefsens:proteins:2020} to robotics \citep{scannellTrajectory2021}.
We have here focused on the Fisher-Rao metric, but other geometries over distributions may apply equally well, e.g.\@ the Wasserstein geometry may be interesting to explore.

\paragraph{Limitations.} The largest practical hurdle with the proposed methodology, is that it only works well for decoders with well-calibrated uncertainties. That is, the decoder should yield high entropy in regions of little training data to ensure that shortest paths follow the trend of the data. This constraint is shared with existing approaches \citep{arvanitidis:iclr:2018}. Some heuristics exists \citep{detlefsen:2019:reliable}, but principled approaches are currently lacking.

\subsubsection*{Acknowledgements}
This work was funded in part by the Novo Nordisk Foundation through the Center for Basic Machine Learning Research in Life Science (NNF20OC0062606). It also received funding from the European Research Council (ERC) under the European Union’s Horizon 2020 research, innovation programme (757360) and from a research grant (15334) from VILLUM FONDEN.

\bibliography{references}
\bibliographystyle{abbrvnat}

\clearpage
\appendix

\thispagestyle{empty}

\onecolumn \makesupplementtitle




\section{Additional details for information geometry}

In this section we provide additional information regarding information geometry. We note that many of these proposition are already know in the literature, however, we include them for completion and for the paper to be standalone.

The Fisher-Rao metric is positive definite only if it is non-singular, and then, defines a Riemannian metric \citep{nielsen:2020}. In this paper, we assume that the observation $\x\in\mathcal{X}$ is a random variable following a probability distribution $p(\x)$ such that $\x\sim p(\x|\eta)$, and any smooth changes of the parameter $\eta$ would alter the observation $\x$. This way, the Fisher-Rao metric used in our paper is non-singular and the statistical manifold $\mathcal{H}$ is a Riemannian manifold.

A known result in \emph{information geometry} \citep{nielsen:2020, amari:2016} is that the Fisher-Rao metric is the first order approximation of the KL-divergence, as recall in Proposition \ref{prop:FRapproxKL}. Using this fact, we can define the Fisher-Rao distance and energy in function of the KL-divergence, leading to Proposition \ref{prop:FRenergydistance}.

\begin{proposition}\label{prop:FRapproxKL}
The Fisher-Rao metric is the first order approximation of the KL-divergence between perturbed distributions: 
\begin{equation}
\mathrm{KL}(p(\x|\eta),~p(\x|\eta+\delta\eta)) = \frac{1}{2} \delta\eta^{\top} \mathbf{I}_\HH(\eta) \delta\eta + o(\delta\eta^2),\nonumber
\end{equation}
with $\mathbf{I}_\HH(\eta) = \int p(\x|\eta) \left[\nabla_\eta\log p(\x|\eta) \nabla_\eta\log p(\x|\eta)\T\right] \dif\x.$
\end{proposition}

\begin{proof}
Let's decompose $\log p(\x|\eta+\delta\eta)$ using the Taylor expansion: 
\[\log p(\x|\eta+\delta\eta) = \log p(\x|\eta) + \nabla_\eta \log p(\x|\eta)\T \delta\eta + \frac{1}{2} \delta\eta\T \Hess_\eta \left[\log p(\x|\eta)\right] \delta\eta + o(\delta\eta^2),\]
where the Hessian is $\Hess_\eta \left[\log p(\x|\eta)\right] = \frac{\Hess_\eta [p(\x|\eta)]}{p(\x|\eta)} - \nabla_\eta\log p(\x|\eta) \nabla_\eta\log p(\x|\eta)\T$ and the $\nabla_\eta \log p(\x|\eta) = \frac{\nabla_\eta p(\x|\eta)}{p(\x|\eta)}$.

Also $\int \nabla_\eta p(\x|\eta) \dif \x = \nabla_\eta\int p(\x|\eta) \dif\x = 0$ and $\int \Hess_\eta[p(\x|\eta)] \dif\x = \Hess_\eta[\int p(\x|\eta) \dif \x] = 0.$

Replacing all those expressions to the first equation finally gives: 
\begin{align*}
     \mathrm{KL}(p(\x|\eta),&~p(\x|\eta+\delta\eta)) = \int p(\x|\eta) \log p(\x|\eta) \dif\x - \int p(\x|\eta) \log p(\x|\eta+\delta\eta) \dif\x \\
    {} & = - \int p(\x|\eta) \left( \nabla_\eta \log p(\x|\eta)\T \delta\eta + \frac{1}{2} \delta\eta\T \Hess_\eta [\log p(\x|\eta)] \delta\eta + o(\delta\eta^2) \right) \dif\x \\
    {}& = \frac{1}{2} \delta\eta\T \left[ \int p(\x|\eta) \left[\nabla_\eta\log p(\x|\eta) \nabla_\eta\log p(\x|\eta)\T\right] \dif\x \right] \delta\eta + o(\delta\eta^2).  
\end{align*}

\end{proof}

\begin{definition}
We consider a curve $\gamma(t)$ and its derivative $\dot{\gamma}(t)$ on the statistical manifold such that, $\forall t\in[0,1], \gamma(t)=\eta_t\in\mathcal{H}$. The manifold is equipped with the Fisher-Rao metric. The length and the energy functionals are defined with respect to the metric $\mathbf{I}_\HH(\eta)$:
\[ \mathrm{Length}(\gamma) = \int_{0}^{1} \sqrt{\dot{\gamma}(t)\T \mathbf{I}_\HH(\eta) \dot{\gamma}(t)} \dif t \quad\text{and}\quad \mathrm{Energy}(\gamma) = \int_{0}^{1} \dot{\gamma}(t)\T \mathbf{I}_\HH(\eta) \dot{\gamma}(t) \dif t.\]

Locally length-minimising curves between two connecting points are called geodesics. These can be found by minimizing the energy using the Euler-Lagrange equations which gives the following system of 2\textsuperscript{nd} order nonlinear ordinary differential equations (ODEs) \citep{arvanitidis:iclr:2018}
\begin{equation}
\label{eq:ode_system}
    \ddot{\gamma}(t) 
    =-\frac{1}{2}\mathbf{I}_{\HH}^{-1}(\gamma(t))\Big[2 (\dot{\gamma}(t)\T \otimes \mathbb{I}_d) \parder{\vectorize{\mathbf{I}_{\HH}(\gamma(t))}}{\gamma(t)}\dot{\gamma}(t)
    - \parder{\vectorize{\mathbf{I}_{\HH}(\gamma(t))}}{\gamma(t)}\T (\dot{\gamma}(t) \otimes \dot{\gamma}(t))\Big].
\end{equation}
\end{definition}

\begin{proposition} \label{prop:FRenergydistance}
The KL-divergence between two close elements of the curve $\gamma$ is defined as: $\mathrm{KL}(p_{t},~p_{t+\delta t})=\mathrm{KL}(p(\x|\gamma(t)),~p(\x|\gamma(t+\delta t)))$. 
The length and the energy functionals can be approximated with respect to this KL-divergence: 
\[\mathrm{Length}(\gamma) \approx\sqrt{2\textstyle\sum_{t=1}^{T} \mathrm{KL}(p_{t},p_{t+\delta t})}  \quad\text{and}\quad \mathrm{Energy}(\gamma) \approx \frac{2}{\delta t} \textstyle\sum_{t=1}^{T} \mathrm{KL}(p_{t},p_{t+\delta t})\]

\end{proposition}

\begin{proof}
On the statistical manifold, we have $\gamma(t+\delta t)=\gamma(t)+\delta t\dot{\gamma}(t)$. The KL-divergence between perturbed distributions can be defined as: $\mathrm{KL}(p_{t},p_{t+\delta t}) = \mathrm{KL}(p(\x|\gamma(t)),p(\x|\gamma(t+\delta t)))= \mathrm{KL}(p(\x|\eta_t),p(\x|\eta_t+\delta\eta_t)),$ with $\eta_t=\gamma(t)$ and $\delta\eta_t=\delta t\ \dot{\gamma}(t)$.
Then, we obtain: 
\[\mathrm{KL}(p_{t},p_{t+\delta t}) = \frac{1}{2} \delta t^2\ \dot{\gamma}(t)\T\mathbf{I}_\HH(\eta_t)\dot{\gamma}(t) + o(\delta t^2).\]

The length and energy terms appear in the following equations:
\begin{align*}
    \int_0^{1}\mathrm{KL}(p_{t},p_{t+\delta t}) \dif t &= \frac{\delta t^2}{2} \int_0^{1} \dot{\gamma}(t)\T\mathbf{I}_\HH(\eta_t)\dot{\gamma}(t) \dif t + o(\delta t^2) = \frac{\delta t^2}{2}\ \mathrm{Energy}(\gamma)+ o(\delta t^2),\\
    \int_0^{1}\sqrt{\mathrm{KL}(p_{t},p_{t+\delta t})} \dif t &= \frac{\delta t}{\sqrt{2}} \int_0^{1} \sqrt{\dot{\gamma}(t)\T \mathbf{I}_\HH(\eta_t)\dot{\gamma}(t)} \dif t + o(\delta t^2) = \frac{\delta t}{\sqrt{2}}\ \mathrm{Length}(\gamma)+ o(\delta t^2).\\
\end{align*}

If we want approximate any continuous function $f$ with a discrete sequence, by partitioning it in $T$ small segments, such that: $\delta t \approx \frac{1}{T}$, we have: $\int_0^{1} f(t) dt \approx \sum_{t=1}^{T} f(t) \delta t$, which in our case gives: 

\[\mathrm{Length}(\gamma) \approx \sqrt{2\textstyle\sum_{t=1}^{T} \mathrm{KL}(p_{t},p_{t+\delta t})} \quad \text{and} \quad \mathrm{Energy}(\gamma) \approx \frac{2}{\delta t} \textstyle \sum_{t=1}^{T} \mathrm{KL}(p_{t},p_{t+\delta t}). \]

\end{proof}

\subsection{The Fisher-Rao metric for several distributions} \label{appendix:FR:distributions}

\begin{table}[h]
\begin{center}
\begin{tabular}{llll}
\hline
Distributions&Probability density functions&Parameters& Fisher-Rao matrix\\[0.2cm] \hline
Normal& $\frac{1}{\sqrt{2\pi\sigma^2}} \exp{-\frac{(\x-\mu)^2}{2\sigma^2}}$ & $\mu, \sigma^2$& $\mathbf{I}_{\mathcal{N}}(\mu,\sigma^2)$\\[0.3cm]
Bernoulli & $\theta^{\x}(1-\theta)^{1-\x}$& $\theta$& $\mathbf{I}_{\mathcal{B}}(\theta)$\\[0.3cm]
Categorical& $\prod_{k=1}^{K}\theta_{k}^{\x_k}$& $\theta_1, \dots, \theta_K$ & $\mathbf{I}_{\mathcal{C}}(\theta_1, \dots, \theta_K)$ \\[0.3cm]
Gamma& $\frac{\beta^{\alpha} \x^{\alpha-1}e^{-\beta \x}}{\Gamma(\alpha)}$    & $\alpha, \beta$& $\mathbf{I}_{\mathcal{G}}(\alpha,\beta)$\\[0.3cm]
Von Mises-Fisher, for $\mathbb{S}^2$ & $\frac{\kappa}{4\pi\sinh{\kappa}}\exp{(\kappa \mu\T \x)}$& $\kappa, \mu$& $\mathbf{I}_{\mathcal{S}}(\kappa,\mu)$\\[0.3cm]
Beta & $\frac{\Gamma(\alpha)\Gamma(\beta)}{\Gamma(\alpha + \beta)} \x^{\alpha - 1}(1-\x)^{\beta -1}$& $\alpha, \beta$ & $\mathbf{I}_{\mathcal{B}}(\alpha,\beta)$ \\[0.3cm] \hline
\end{tabular}
\end{center}
\caption{List of distributions}
\label{tab:distributions}
\end{table}

With the notations of Table \ref{tab:distributions}, the Fisher-Rao matrices of the the univariate Normal, Bernoulli and Categorical are: 
\[
\mathbf{I}_{\mathcal{N}}(\mu,\sigma^2) = \begin{pmatrix} \frac{1}{\sigma^2} & 0 \\ 0 & \frac{1}{2\sigma^2}\\\end{pmatrix}, \quad
\mathbf{I}_{\mathcal{B}}(\theta) = \frac{1}{\theta(1-\theta)}, \quad
\mathbf{I}_{\mathcal{C}}(\theta_1, \dots, \theta_K) = \begin{pmatrix}
  \sfrac{1}{\theta_1} & 0 &\dots & 0 \\ 
  0 & \sfrac{1}{\theta_2} & \dots & 0 \\ 
  \vdots & \vdots & \ddots & \vdots \\ 
  0 & 0 & \dots & \sfrac{1}{\theta_K} \\ 
\end{pmatrix}
\]
In addition, the Fisher-Rao matrices of the Gamma, Von Mises-Fisher and the Beta distributions are:
\begin{align*}
    \mathbf{I}_{\mathcal{G}}(\alpha,\beta) &= \begin{pmatrix} \frac{\alpha}{\beta^2} & -\frac{1}{\beta} \\ -\frac{1}{\beta} & \Psi_1(\alpha) \\\end{pmatrix}, \\
    \mathbf{I}_{\mathcal{S}}(\kappa,\mu) &= \begin{pmatrix} \kappa K(\kappa)(\mathds{1}-3\mu\mu^{\top})+\kappa^2\mu\mu^{\top} & (\kappa K(\kappa)^2-\frac{2}{k}K(\kappa)+1)\mu \\ (\kappa K(\kappa)^2-\frac{2}{k}K(\kappa)+1)\mu^{\top} & 3K(\kappa)^2-\frac{2}{\kappa}K(\kappa)+1 \\\end{pmatrix}, \\
    \mathbf{I}_{\mathcal{B}}(\alpha,\beta) &= \begin{pmatrix} \Psi_1(\alpha)-\Psi_1(\alpha+\beta) & -\Psi_1(\alpha+\beta) \\ -\Psi_1(\alpha+\beta) & \Psi_1(\beta) - \Psi_1(\alpha+\beta) \\\end{pmatrix},
\end{align*}
with $\Psi_1(\alpha) = \frac{\partial^2 \ln\Gamma(\alpha)}{\partial \alpha}$ the trigamma function, and $K(\kappa) = \coth{\kappa}-\frac{1}{\kappa}$.

\begin{proof}
The univariate Normal, Bernoulli and Categorical have already been studied by \cite{tomczak:2012}, and the Beta distribution by \cite{brigant:2019}. We will then focus our proof on the Gamma and the Von-Mises Fisher distributions. 

In order to bypass unnecessary details, we will use the following notations, we redefine the Fisher-Rao as: $\mathbf{I}(\eta)=\mathbb{E}_{x}[g(\eta, x)g(\eta, x)\T]$, with $g(\eta,x) = \nabla_{\eta} \ln p(x|\eta)$ the Fisher score. We call $G = g(\eta, x)g(\eta, x)^{\top}$, and $G_{ij}$ the matrix elements. 

\textbf{Gamma distribution:}

We have $p(x|\alpha, \beta) = \Gamma(\alpha)^{-1} \beta^{\alpha} x^{\alpha-1}e^{-\beta x}$, which leads to: 
\begin{align*}
    \ln p(x|\alpha, \beta) &= - \ln \Gamma(\alpha) + \alpha \ln \beta + (\alpha - 1) \ln x - \beta x, \\
    \frac{\partial \ln p}{\partial \alpha} & = - \Psi(\alpha) + \ln \beta + \ln x, \\
    \frac{\partial \ln p}{\partial \beta} & = \frac{\alpha}{\beta} - x. 
\end{align*}
Then:
\begin{align*}
    G_{11} &= \left(\frac{\partial \ln p}{\partial \alpha}\right)^2 = (\Psi_{0}(\alpha) + \ln \beta)^2 + 2 (\Psi(\alpha) + \ln \beta ) ln x + \ln^2 x, \\
    G_{22} &= \left(\frac{\partial \ln p}{\partial \beta}\right)^2  = \left(\frac{\alpha}{\beta}\right)^2 - 2 \frac{\alpha}{\beta} x + x^2,  \\
    G_{12} = G_{21} &= \frac{\partial \ln p}{\partial \alpha} \cdot \frac{\partial \ln p}{\partial \beta}  = \left(\Psi(\alpha) + \ln \beta\right) \left(\frac{\alpha}{\beta} - x\right) + \frac{\alpha}{\beta} \ln x - x \ln x.
\end{align*}

We know that $\mathbb{E}[x] = \frac{\alpha}{\beta}$. We can compute, using your favorite symbolic computation software, the following moments: 
\begin{align*}
    \mathbb{E}[\ln x] &= -\ln \beta + \Psi(\alpha) \\
    \mathbb{E}[x \ln x] &= \frac{\alpha}{\beta} \left(\Psi(\alpha + 1) - \ln \beta\right) \\
    \mathbb{E}[\ln^2 x] &= \left(\ln \beta - \Psi(\alpha)\right)^2 + \Psi_1(\alpha)
\end{align*}
Replacing the moments for the following equations: $\mathbb{E}[G_{11}]$, $\mathbb{E}[G_{22}]$ and $\mathbb{E}[G_{12}]$ will finally give the Fisher-Rao matrix.\\

\textbf{Von Mises Fisher distribution, for $\mathbb{S}^2$:}

We have $p(\x|\mu, \kappa) = C_{3}(\kappa) \exp (\kappa \mu\T \x)$, with $C_{3}(\kappa) = \kappa (4 \pi \sinh \kappa)^{-1}$. Here, $\mu$ is a 3-dimensional vector with $\lVert \mu \rVert = 1$. 
\begin{align*}
    \ln p(\x|\mu, \kappa) &= \ln \kappa - \ln 4\pi - \ln \sinh(\kappa) + \kappa \mu\T \x \\
    \nabla_{\mu} \ln p & = \kappa \x  \\
    \frac{\partial \ln p}{\partial \kappa} & =  \kappa^{-1} - \text{coth}(\kappa) + \mu\T \x.
\end{align*}
Here, the Fisher-Rao matrix $\mathbf{I}_{\mathcal{S}}$ will be composed of block matrices, such that: $\mathbf{I}_{\mathcal{S}} = \mathbb{E}[G]$, with $G_{11}$ a $3\times 3$-matrix, $G_{22}$ a scalar, and $G_{12}=G_{21}^{\top}$ a 3-dimensional vector. 
\begin{align*}
    G_{11} &= \nabla_{\mu}\ln p \nabla_{\mu}\ln p^{\top} = \kappa^2 \x \x\T \\
    G_{22} &= \left(\frac{\partial \ln p}{\partial \kappa}\right)^2 = K(\kappa)^2 + 2 K(\kappa) \mu^{\top} \x + (\mu\T \x)^2  \\
    G_{12} = G_{21}^{\top} &= \frac{\partial \ln p}{\partial \kappa} \cdot \nabla_{\mu}\ln p = \left(K(\kappa) + \mu\T \x \right) \kappa \x, 
\end{align*}
with $K(\kappa) = \text{coth}(\kappa) - \frac{1}{\kappa}$.

We know from \cite{hillen:2016} that the mean and variance of the Von Mises Fisher distribution in the 3-dimensional case is: $\mathbb{E}[\x] = K(\kappa) \mu$ and $\text{Var}[\x] = \frac{1}{\kappa} K(\kappa) \mathds{1} + (1- \frac{\text{coth}(\kappa)}{\kappa} + \frac{2}{\kappa^2} - \text{coth}^2(\kappa)) \mu \mu^{\top}$. We can then deduce the following meaningful moments: 
\begin{align*}
    \mathbb{E}[\x\x^{\top}] & = \text{Var}[\x] + \mathbb{E}[\x]\mathbb{E}[\x]^{\top} = \left(1-\frac{3}{\kappa} K(\kappa)\right) \mu\mu^{\top} + \frac{1}{\kappa} K(\kappa) \mathds{1}, \\
    \mathbb{E}[\mu\T \x] & = \mu^{\top} \mathbb{E}[\x] = K(\kappa) \mu^{\top} \mu = K(\kappa) \\
    \mathbb{E}[(\mu\T \x)^2] & = \mu^{\top} \text{Var}[\x] \mu + \mathbb{E}[\mu\T \x]^2 = 1-\frac{2}{\kappa}K(\kappa), \\
    \mathbb{E}[\mu\T \x \x] &= \mathbb{E}[\mu \x \x\T] = \mathbb{E}[\x \x\T] \mu = \left(1 - \frac{3}{\kappa} K(\kappa)\right) \mu + \frac{1}{\kappa} K(\kappa) \mu.
\end{align*}
Replacing those moments in the following expressions: $\mathbb{E}[G_{11}]$, $\mathbb{E}[G_{22}]$, $\mathbb{E}[G_{12}]$ directly gives the Fisher-Rao metric.
\end{proof}

\section{Curve energy approximation for categorical data}
\label{appendix:categorical_details}

In this section we present the details of the example in Section~\ref{sec:example_categorical}. In particular, we the steps to derive an approximation to the energy of a latent curve in closed form, which is suitable for applying automatic differentiation. This is particularly useful for our setting, since it allows us to consider our framework as a Black Box Random Geometry processing toolbox.

Let a random variable $\x\in\R^D$ that follows a generalized Bernoulli likelihood $p(\x|\eta)$, so the vector $\x\in\R^D$ is of the form $\b{x} = (0, \cdots, 1, \cdots, 0)$ with $\sum_i x_i = 1$. The parameters $\eta\in\R^D$ are given as $\eta = h(\z)$, with $\eta_i \geq 0~\forall i$ and $\sum_i \eta_i = 1$ so we know that the parameters lie on the unit simplex. Actually, they represent the probability the corresponding dimension to be 1 on a random draw. Also, the $p(\x|\z) = \eta_1^{[x_1]}\cdots \eta_D^{[x_D]}$, where $[x_i] = 1$ if $x_i=1$ else $[x_i] = 0$ which can be seen as an indicator function. The $\log p(\x|\eta) = \sum_i [x_i] \log(\eta_i)$ and $\nabla_\eta \log p(\x|\eta) = \left(\frac{[x_1]}{\eta_1}, \dots, \frac{[x_D]}{\eta_D}\right)$. Due to the outer product we have to compute the following expectations
\begin{align}
    &\mathbb{E}_\b{x}\left[ \frac{[x_i]}{\eta_i} \frac{[x_j]}{\eta_j}\right] = 0, ~\quad \text{if}\quad i\neq j,\\
    &\mathbb{E}_\b{x}\left[ \left(\frac{[x_i]}{\eta_i}\right)^2\right] = \frac{1}{\eta_i}, ~\quad \text{if}\quad i=j,
\end{align}
because the $[x_i]$ and $[x_j]$ cannot be 1 on the same time, while the $\mathbb{E}_\b{x}[[x_i]^2] = \eta_i$ as it shows the number of times $x_i=1$. So the Fisher-Rao metric of $\HH$ is equal to $\b{I}_\HH(\eta) = \mathrm{diag}\left(\sfrac{1}{\eta_1},\dots,\sfrac{1}{\eta_D}\right)$. Note that the shortest paths between two distributions must be on the unit simplex in $\HH$, while on the same time respecting the geometry of the Fisher-Rao metric.

We can easily parametrize the unit simplex by $[\eta_1,\dots,\eta_{D-1}, \widetilde{\eta}_D]$ with
\begin{equation}
    \widetilde{\eta}_D(\eta_1,\dots, \eta_{D-1}) = 1 - \sum_{i=1}^{D-1} \eta_i.
\end{equation}
This allows to pullback the Fisher-Rao metric in the latent space $[\eta_1,\dots,\eta_{D-1}]$ as we have described in this paper. Intuitively, the $\z = [\eta_1,\dots,\eta_{D-1}]$ and the function $h$ is the parametrization of the simplex. Hence, we are able to compute the shortest path using the induced metric.

However, there is a simpler way to compute this path. We know that the element-wise square root of the parameters $\eta$ gives a point on the positive orthonant of the unit sphere as $y_i = \sqrt{\eta_i}~\Rightarrow~\sum_i y_i^2 = \sum_i \sqrt{\eta_i}^2 = 1$. We also know that the shortest path on a sphere is the great-circle. Therefore, the distance between two distributions parametrized by $\eta$ and $\eta'$ on the unit simplex in $\HH$, can be equivalently measured using the great-circle distance between their square roots as
\begin{equation}
    \mathrm{dist}(\eta, \eta') = \arccos\sqrt{\eta}\T\sqrt{\eta'}.
\end{equation}

In this way, we can approximate the energy of a curve $c(t)$ in the latent space as follows
\begin{align}
\label{eq:approx_energy_categorical}
    \mathrm{Energy}[c]&\approx \sum_{n=1}^{N-1}\mathrm{dist}^2(h(c(\sfrac{n}{N})),h(c(\sfrac{n+1}{N}))) = \sum_{n=1}^{N-1}\arccos^2 \sqrt{h(c(\sfrac{n}{N}))}\T\sqrt{h(c(\sfrac{n+1}{N}))}\nonumber\\
    &=\sum_{n=1}^{N-1}\left(2 - 2\sqrt{h(c(\sfrac{n}{N}))}\T\sqrt{h(c(\sfrac{n+1}{N}))}\right),
\end{align}
where we used at the last step the small angle approximation $\cos\theta \approx 1 - \frac{\theta^2}{2} \Leftrightarrow \theta^2 \approx 2 - 2\cos\theta$. Note that this formulation is suitable for our proposed method to compute shortest paths (see Section~\ref{sec:geodesic_computation}).

The derivation above represents the conceptual strategy, while in general we proposed to use the KL divergence approximation result \eqref{eq:KL_approx_inner_product} in place of the great-circle distance. Intuitively, when the KL divergence has an analytic solution, we can derive an analogous energy approximation. Even if the solution of the KL is intractable, we can still use our approach as long as we can estimate the KL using Monte Carlo and propagate the gradient through the samples using a re-parametrization scheme or a score function estimator.

\section{Information geometry in generative modeling}

In this section we present the additional technical information related to the pullback Fisher-Rao metric in the latent space of a VAE.

\subsection{Details for the pullback metric in the latent space}

We call $h$ the non linear function, typically parametrized as deep neural networks, that maps the variables from the latent space $\mathcal{Z}$ to the parameter space $\mathcal{H}$, such that: $h(\z)=\eta$, with $\z\in\mathcal{Z}$ and $\eta\in\mathcal{H}$. Furthermore, the data $\x \in \mathcal{X}$ is reconstructed such that it follows a specific distribution: $\x \sim p(\x|\eta)$, with $p(\x|\eta)$ being for instance a Bernoulli or Gaussian distribution. The parameter space $\mathcal{H}$ is a statistical manifold equipped with Fisher-Rao metric: $\b{I}_\HH(\eta) \overset{\Delta}{=} \int p(\x|\eta) \left[\nabla_\eta\log p(\x|\eta) \nabla_\eta\log p(\x|\eta)\T\right] \dif \x$. We denote by $\b{J}_h$ the Jacobian of $h$.

\begin{proposition} \label{prop:riemannian-pullback}
The latent space $\mathcal{Z}$ is equipped with the Riemannian pullback metric tensor: 
\[ \b{M}(\z) \overset{\Delta}{=} \b{J}_{h}(\z)\T \b{I}_\HH(h(\z)) \b{J}_{h}(\z). \]
\end{proposition}

\begin{proof}
The parameter space is a statistical manifold equipped with the Fisher-Rao metric $\b{I}_\HH(\eta)$, thus the scalar product at $\eta$ between two vectors $d\eta_1, d\eta_2 \in \mathcal{H}$ is: $\langle d\eta_1, d\eta_2 \rangle_{\b{I}_\HH(\eta)} =  d\eta_1\T \b{I}_\HH (\eta) d\eta_2$.  For two vectors $d\z_1, d\z_2 \in\mathcal{Z}$, we have at $\eta = f(\z)$ that: $\langle d\eta_1, d\eta_2 \rangle_{\b{I}_\HH(\eta)} =  \langle \b{J}_h(\z) d\z_1, \b{J}_h(\z) d\z_2 \rangle_{\b{I}_\HH(\eta)} = d\z_1\T (\b{J}_{h}(\z)\T \b{I}_\HH(h(\z)) \b{J}_{h}(\z)) d\z_2$. 

$\b{I}_\HH(h(\z))$ is a Riemannian metric tensor by definition, and it is then positive definite. Furthermore, $h:\mathcal{Z}\to\mathcal{H}$ is a smooth immersion, and so $\b{J}_{h}(\z)$ is full-rank. It follows that $\b{J}_{h}(\z)\T \b{I}_\HH(h(\z)) \b{J}_{h}(\z)$ is positive definite. Hence $\b{M}(\z)$ is a Riemannian metric tensor.
\end{proof}


\begin{proposition} \label{prop:pullback-FR-equal}
Our pullback metric $\b{M}(\z)$ is actually equal to the Fisher-Rao metric obtained over the parameter space $\Z$: 
\[\b{M}(\z) = \b{I}_\Z(\z) \overset{\Delta}{=} \int p(\x|\z) \left[\nabla_\z \log p(\x|\z) \nabla_\z \log p(\x|\z)\T\right] \dif \x \]
\end{proposition}


\begin{proof}
We will show that $\b{I}_\Z(\z) = \b{J}_{f}(\z)\T \b{I}_\HH(\eta) \b{J}_{f}(\z)$. 
Let's consider the definition of the Fisher-Rao metric in $\Z$ :
\begin{align}
    \b{I}_\Z(\b{z}) 
    &= \int \nabla_{\b{z}} \log p(\b{x}~|~\b{z}) \cdot \nabla_{\b{z}} \log p(\b{x}~|~\b{z})\T p(\b{x}~|~\b{z})\dif \b{x}\\
    &= \int \b{J}_f(\b{z})\T \nabla_{\eta} \log p(\b{x}|\eta)\nabla_{\eta} \log p(\b{x}|\eta)\T \b{J}_f(\b{z}) p(\b{x}|\eta)\dif\b{x} \\
    &= \b{J}_f(\b{z})\T \left[ \int_\X \nabla_{\eta} \log p(\b{x}|\eta)\nabla_{\eta} \log p(\b{x}|\eta)\T  p(\b{x}|\eta)\dif\b{x} \right]\b{J}_f(\b{z})\\
    &= \b{J}_f(\b{z})\T \b{I}_\HH(f(\z)) \b{J}_f(\b{z})=\b{M}(\z) \nonumber
\end{align}
where we use the fact that $\eta=f(\b{z})$ so the $\nabla_{\b{z}}\log p(\b{x}|f(\b{z})) = \b{J}_f(\b{z})\T \cdot \nabla_{\eta} \log p(\b{x}|\eta)$

The same argument can be proved as follows:
\begin{align}
    &\inner{d\eta}{\b{I}_\HH(\eta) d\eta} = \inner{\b{J}_f(\b{z})d\b{z}}{\b{I}_\HH(f(\b{z}))\b{J}_f(\b{z})d\b{z}} \\
    &= \inner{\b{J}_f(\b{z})d\b{z}}{\int \nabla_\eta \log p(\b{x}|\eta) \nabla_\eta \log p(\b{x}|\eta) \T p(\b{x}|\eta)\dif\b{x} ~\b{J}_f(\b{z})d\b{z}}\\
    &=\inner{d\b{z}}{\int \b{J}_f(\b{z})\T \cdot \nabla_\eta \log p(\b{x}|\eta) ~\nabla_\eta \log p(\b{x}|\eta)\T \cdot \b{J}_f(\b{z}) p(\b{x}|\eta)\dif\b{x} ~d\b{z}}\\
    &=\inner{d\b{z}}{\int \nabla_{\b{z}}\log p(\b{x}|\b{z}) \nabla_{\b{z}}\log p(\b{x}|\b{z})\T p(\b{x}|\b{z}) \dif\b{x}~d\b{z}} = \inner{d\b{z}}{\b{I}_\Z(\b{z}) d\b{z}}
\end{align}
\end{proof}

In section \ref{appendix:FR:distributions}, we have seen how to derive a close-form expression of the Fisher-Rao metric for a one-dimensional observation $x$ that follows a specific distribution. In practice, $\x\in\mathcal{X}\subset\mathbb{R}^D$ is a multi-dimensional variable where each dimension represents, for instance, a pixel when working with images or a feature when working with tabular data. Each feature, $x_i$ with $i=1\cdots D$, is obtained for a specific set of parameters $\{\eta_i\}$. We assume that the features follow the same distribution $\mathcal{D}$, such that: $x_i \sim p(x_i|\eta_i)$, and $p(\x|\eta) = \prod_{i=1}^{D}p(x_i|\eta_i)$. 

\begin{proposition} \label{prop:blockmatrix}
If the features follow the same distribution $\mathcal{D}$, such that: $x_i \sim p(x_i|\eta_i)$ and $p(\x|\eta) = \prod_{i=1}^{D}p(x_i|\eta_i)$, then the Fisher-Rao metric $\mathbf{I}_{\HH}(\eta)$ is a block matrix where the diagonal terms are the Fisher-Rao matrices $\mathbf{I}_{\HH,i}$ obtained for each data feature $x_i$: 
\[
\mathbf{I}_{\HH}(\eta) = \begin{pmatrix}
  \b{I}_{\HH,1} & 0 &\dots & 0 \\ 
  0 & \b{I}_{\HH,2} & \dots & 0 \\ 
  \vdots & \vdots & \ddots & \vdots \\ 
  0 & 0 & \dots & \b{I}_{\HH, D} \\ 
\end{pmatrix}
\]
\end{proposition}
\begin{proof}
We have $x_i\sim p(x_i|\eta_i)$ and  $\mathbf{I}_{\HH,i}=\int p(x_i|\eta_i) \left[\nabla_{\eta_i}\log p(x_i|\eta_i) \nabla_{\eta_i}\log p(x_i|\eta_i)\T\right] dx_i$. 
Also, we assumed: $p(\x|\eta)=\prod_{i=1}^{D}p(x_i|\eta_i)$.
We then have: $\log{p(\x|\eta)} = \sum_{i=1}^{D} \log{p(x_i|\eta_i)}$, and the Fisher score: $\nabla_{\eta}\log{p(\x|\eta)} = \nabla_{\eta} \sum_{i=1}^{D} \log{p(x_i|\eta_i)} = \left[\nabla\eta_1 \ln{p(x_1|\eta_1)}, \dots, \nabla\eta_D \ln{p(x_1|\eta_D)}\right]\T$. \\

\noindent
The matrix $\mathbf{I}_{\HH}(\eta)$ is thus a $D \times D$ block matrix, where the $(i,j)$-block element is: 
\[I_{ij}=\int p(x_i|\eta_i) \left[\nabla_{\eta_i} \log p(x_i|\eta_i) \nabla_{\eta_i}\log p(x_j|\eta_j)\T\right] dx_i.\] 
Let's note that: 
\[\int p(x_i|\eta_i) \nabla_{\eta_i}\log p(x_i|\eta_i) dx_i = \int p(x_i|\eta_i) \frac{\nabla_{\eta_i} p(x_i|\eta_i)}{p(x_i|\eta_i)}dx_i = \nabla_{\eta_i} \int p(x_i|\eta_i) dx_i = 0.\]
When $i=j$, we have $\mathbf{I}_{ii}=\mathbf{I}_{\HH,i}$, with $\mathbf{I}_{\HH,i}$ being the Fisher-Rao metric obtained for: $x_i\sim p(x_i|\eta_i)$. \\
\noindent
When $i\neq j$, we have: $\mathbf{I}_{ij}=\nabla\log p(x_j|\eta_j)\T \int p(x_i|\eta_i) \nabla_{\eta_i}\log p(x_i|\eta_i) dx_i = 0$.  
\end{proof}

Then, for example, if we are dealing with binary images, and make the assumption that each pixel $x_i$ follows a Bernoulli distribution: $p(x_i|\eta_i) = \eta^{x_i}(1-\eta_i)^{1-x_i}$, then according to Section \ref{appendix:FR:distributions} and Proposition \ref{prop:blockmatrix}, the Fisher-Rao matrix that endows the parameter space $\mathcal{H}$ is:
\[
\b{I}_\HH(\eta) = \begin{pmatrix}
              \frac{1}{\eta_1(1-\eta_1)} & 0 &\dots & 0 \\ 
              0 & \frac{1}{\eta_2(1-\eta_2)} & \dots & 0 \\ 
              \vdots & \vdots & \ddots & \vdots \\ 
              0 & 0 & \dots & \frac{1}{\eta_D(1-\eta_D)} \\ 
            \end{pmatrix}.
\]

We have seen that in theory, we can obtain a close form expression for the pullback metric, if the probability distribution is known. In practice, we can directly infer the metric using the approximation of the KL-divergence. 

\begin{proposition}
\label{appendix:prop:metric_approximation}
We define perturbations vectors as: $\delta e_{i} = \varepsilon \cdot \b{e_i}$, with $\varepsilon\in\mathbb{R}_{+}$ a small infinitesimal quantity, and ($\b{e_i}$) a canonical basis vector in $\mathbb{R}^d$. For better clarity, we rename $\mathrm{KL}(p(\x|\z), p(\x|\z+\delta \z)) = \mathrm{KL}_{\z}(\delta \z)$ and we note $\b{M}_{ij}$ the components of $\b{M}(\z)$.
We can then approximate by a system of equations the diagonal and non-diagonal elements of the metric:
    \begin{align*}
        \b{M}_{ii} &\approx 2 \ \mathrm{KL}_{\z}(\delta \b{e_i})/ \varepsilon^2 \ \\
        \b{M}_{ij} = \b{M}_{ji} &\approx  \left(\mathrm{KL}_{\z}(\delta \b{e_i} + \delta \b{e_j})- \mathrm{KL}_{\z}(\delta \b{e_i}) -\mathrm{KL}_{\z}(\delta \b{e_j}) \right)/ \varepsilon^2. 
    \end{align*}
\end{proposition}

\begin{proof}
From Proposition \ref{prop:FRapproxKL}, we know that: 
\[\mathrm{KL}_{\z}(\delta \z) = \frac{1}{2} \delta\z^{\top} \b{M}(\z) \delta\z + o(\delta\z^2).\] 

Let's take $\delta e_{i} = \varepsilon \cdot \b{e_i}$. On one hand, we have: $\delta e_{i}^{\top} \b{M}(\z) \delta e_{i} = \varepsilon^2 \b{M}_{ii}$. On the second hand, we also have: $ \delta e_{i}^{\top} \b{M}(\z) \delta e_{i} \approx 2 \mathrm{KL}_{\z}(\delta \b{e_i})$, which gives us the equation to infer the diagonal elements of the metric. 

Now, let's take $\delta e_{i} + \delta e_j = \varepsilon \cdot (\b{e_i}+\b{e_j})$. Then, we have: $(\delta e_{i} + \delta e_j)^{\top} \b{M}(\z) (\delta e_{i} + \delta e_j) = \varepsilon^2 (\b{M}_{ii} + \b{M}_{jj} + \b{M}_{ij} + \b{M}_{ji})$. We also know that $\b{M}_{ji} =\b{M}_{ij}$. Again, we also have: $(\delta e_{i} + \delta e_j)^{\top} \b{M}(\z) (\delta e_{i} + \delta e_j) \approx 2 \mathrm{KL}_{\z}(\delta \b{e_i} + \delta \b{e_j})$.

We can replace the terms $\b{M}_{ii}$ and $\b{M}_{jj}$ in the equation obtained above with the KL-divergence for the diagonal terms. Which finally gives us: $\b{M}_{ij} = \b{M}_{ji} \approx  \left(\mathrm{KL}_{\z}(\delta \b{e_i} + \delta \b{e_j})- \mathrm{KL}_{\z}(\delta \b{e_i}) -\mathrm{KL}_{\z}(\delta \b{e_j}) \right)/ \varepsilon^2$.

\end{proof}

\subsection{Uncertainty quantification and regularization}
\label{appendix:sec:UQ}

As discussed in the main text, we carefully design our mappings from latent space to parameter space such that they model the training codes according to the learned decoders, and extrapolate to uncertainty outside the support of the data. This, we refer to as \textbf{uncertainty regularization}. In this section we explain it in detail. The core idea of this uncertainty regularization is imposing a ``slider'' that forces the distribution $p(\x|\z)$ to change when $\z$ is far from the training latent codes. For this, we use a combination of KMeans and the sigmoid activation function.

We start by encoding our training data, arriving at a set of latent codes $\{\z_n\}_{n=1}^{N}\subseteq\mathcal{Z}$. We then train KMeans($k$) on these latent codes (where $k$ is a hyperparameter that we tweak manually), arriving at $k$ cluster centers $\{\b{c}_j\}_{j=1}^k$. These cluster centers serve as a proxy for "closeness" to the data: we know that a latent code $\z\in\mathcal{Z}$ is near the support if $D(\z) := \min_j\left\{\|\z - \b{c}_j\|^2\right\}$ is close to 0.

The next step in our regularization process is to reweight our decoded distributions such that we decode to high uncertainty when $D(\z)$ is large, and we decode to our learned distributions when $D(\z) \approx 0$. This mapping from $[0, \infty)\to (0, 1)$ can be constructed using a modified sigmoid function \cite{detlefsens:proteins:2020,detlefsen:2019:reliable}, consider indeed
\begin{equation}
\label{appendix:eq:translated_sigmoid}
\tilde{\sigma}_{\beta}(d) = \text{Sigmoid}\left(\frac{d - c\cdot\text{Softplus}(\beta)}{\text{Softplus}(\beta)}\right),
\end{equation}
where $\beta\in\mathbb{R}$ is another hyperparameter that we manually tweak, and $c\approx 7$.

With this translated sigmoid, we have that $\tilde{\sigma}_\beta(D(\z))$ is close to 0 when $\z$ is close to the support of the data (i.e. close to the cluster centers), and it converges to 1 when $D(\z)\to\infty$.  $\tilde{\sigma}_\beta(D(\z))$ serves, then, as a slider that indicates closeness to the training codes. This reweighting takes the following form:
\begin{equation}
\label{appendix:eq:reweight}
\text{reweight}(\z) = (1 - \tilde{\sigma}_\beta(D(\z)))h(\z) + \tilde{\sigma}_\beta(D(\z))\,\text{extrapolate}(\z),
\end{equation}
where  $h(\z) = \eta\in\mathcal{H}$ represents our learned networks in parameter space,  and $\text{extrapolate}(\z)$ returns the parameters of the distribution that maximize uncertainty (e.g. $\sigma\to\infty$ in the case of an isotropic Gaussian, $p\to 1/2$ in the case of a Bernoulli, and $\kappa\to 0$ in the case of the von Mises-Fisher).

For the particular case of the experiment in which we pull back the Fisher-Rao metric from the parameter space of several distributions (see \ref{sec:toy_latent_space_experiment}), Table \ref{appendix:tab:extrapolation_mechanisms} provides the exact extrapolation mechanisms and implementations of $h(\z)$.


\begin{table}[]
    \centering
    \resizebox{\textwidth}{!}{
    \begin{tabular}{lll}
        Distribution & {$h\colon\mathcal{Z}_{\text{toy}}\to\mathcal{H}$} & Extrapolation mechanism \\
        \toprule
        Normal & $\mu(\z)=10\cdot f_3(\z)$, ~ $\sigma(\z) = 10\cdot \text{Softplus}(f_3(\z))$ & $\sigma(\z)\to\infty$ \\[0.1cm]
            Bernoulli & {$p(\z) = \text{Sigmoid}(f_{15}(\z))$} & $p(\z) = 1/2$ \\[0.1cm]
            Beta & $\alpha(\z)=10\cdot \text{Softplus}(f_3(\z))$, ~  $\beta(\z)=10\cdot \text{Softplus}(f_3(\z))$ & $(\alpha(\z), \beta(\z)) = (1, 1)$ \\[0.1cm]
            Dirichlet & $\alpha(\z)=\text{Softplus}(f_3(\z))$  & $\alpha(\z) = 1$ \\[0.1cm]
            Exponential & $\lambda(\z)=\text{Softplus}(f_3(\z))$  & $\lambda(\z) \to 0$ \\[0.2cm]
    \end{tabular}
    }
    \caption{This table shows the implementations of the decode and extrapolate functions in Eq. (\ref{appendix:eq:reweight}) for all the distributions studied in our second experiment (see Sec. \ref{sec:toy_latent_space_experiment}). Here we represent a randomly initialized neural network with $f_i$, where $i$ represents the size of the co-domain. For example, in the case of the Dirichlet distribution, we use a randomly initialized neural network to compute the parameters $\alpha$ of the distribution and, since these have to be positive, we pass the output of this network through a Softplus activation; moreover, since the Dirichlet distribution is approximately uniform when all its parameters equal 1, our extrapolation mechanism consists of replacing the output of the network with a constant vector of ones.}
    \label{appendix:tab:extrapolation_mechanisms}
\end{table}

\section{Details for our implementation and experiments}

In this section we present the technical details that we used in our implementation and experiments. We are currently implementing an open-source version of our code \href{https://github.com/MachineLearningLifeScience/stochman/tree/black-box-random-geometry/examples/black_box_random_geometries}{\texttt{here}}.

\subsection{What we mean when we say black-box random geometry}
\label{appendix:sec:black_box_random_geometry}

Before we dive into the specific details of our experiments, it is worth noting that they were all made using the same \textit{interface}. This is precisely what we mean when we say that our results open the doors for black-box random geometry: We can define a \texttt{curve\_energy} method that is agnostic to the distribution our models decode to.

To hammer this point home, consider the following interface, written in Python:

\begin{lstlisting}[language=Python]
class StatisticalManifold:
    def __init__(self, model: torch.nn.Module):
        # A model with regularized uncertainty (see Uncertainty Quantification)
        self.model = model
        assert "decode" in dir(model)

    def curve_energy(self, curve: CubicSpline) -> torch.Tensor:
        # An energy function that can be minimized using autodifferentiation.

        dt = (curve[1] - curve[0])
        dist1 = self.model.decode(curve[:-1])
        dist2 = self.model.decode(curve[1:])
        kl = kl_divergence(dist1, dist2)
        energy = kl.sum() * (2 * dt ** -1)

        return energy
\end{lstlisting}

Notice that the user need only provide a \texttt{model} that implements a \texttt{decode} function which is expected to return a distribution with proper uncertainty estimates (as described in Sec.~\ref{appendix:sec:UQ}). Line 14 is a direct implementation of our derived expression for the energy (see Prop. \ref{prop:FRenergydistance}). Most distributions of interest are available in the Torch submodule \texttt{torch.distributions}, and similar implementations could be done for other frameworks.

\subsection{Shortest path approximation with cubic splines}
\label{sec:geodesic_solver_details}

As we described in the main paper, we use an approximate solution for the shortest paths based on cubic splines. Let a cubic spline $c_{\bs{\psi}}(t) = [1,t,t^2,t^3]\T[\bs{\psi}_0, \bs{\psi}_1, \bs{\psi}_2, \bs{\psi}_3]$ with parameters $\bs{\psi}_i\in\R^{d \times 1}$. Also, in our implementation the actual curve is a piecewise cubic spline and we optimize the $K$ control points $\b{c}_k$ as well. We optimize the parameters using the approximation of the curve energy $\{\bs{\psi}_k^*, \b{c}^*_k\}_{k=1}^K =  \argmin_{\bs{\psi}} \mathrm{Energy}[c_{\bs{\psi}}]$. In general, we can use Prop.~\ref{prop:FRenergydistance} as long as we can propagate the gradient through the KL or as in \eqref{eq:approx_energy_categorical} if an explicit closed form solution exists. In this case, we are able to use automatic differentiation for the optimization of the parameters (as discussed in Sec.~\ref{appendix:sec:black_box_random_geometry}).

In practical terms, we compute these shortest paths by creating a uniform grid in latent space and computing, only once, the curve energy for the edges of this grid. After this expensive computation (which only needs to be performed once) we can use shortest-paths algorithms in graphs to create a suitable initialization of the geodesic. We fit a cubic spline to this initialization and then optimize its parameters further. 

\subsection{Models used}

\begin{table}
\centering
\begin{tabular}{@{}rl@{}}
\multicolumn{2}{c}{\textbf{Pulling back the Euclidean vs. Fisher-Rao} (Sec. \ref{sec:compare_metrics_with_gaussians_generators_uq})}\\
\toprule
\textbf{Module}  & \multicolumn{1}{c}{\textbf{MLP}}     \\ \midrule
\multicolumn{2}{c}{Encoder}  \\
$\mu$ & Linear(728, 2) \\
\midrule
\multicolumn{2}{c}{Decoder}  \\
$\mu$ & Linear(2, 728) \\
$\sigma_{\text{UR}}$ & RBF(), PosLinear(500, 1), Reciprocal(), PosLinear(1, 728) \\
$\sigma_{\text{no UR}}$ & Linear(2, 728), Softplus() \\
\midrule
Optimizer  & \multicolumn{1}{@{}l@{}}{Adam ($\alpha = 1 \times 10^{-5}$)}  \\
Batch size             & \multicolumn{1}{@{}l@{}}{32} \\
\bottomrule
\end{tabular}
\vspace{0.2cm}
\caption{ This table shows the Variational Autoencoder used in our first experiment (see Sec. \ref{sec:compare_metrics_with_gaussians_generators_uq}). The network for approximating the standard deviation $\sigma$ leverages ideas from \cite{arvanitidis:iclr:2018}, in which an RBF network is trained on latent codes using centers positioned through KMeans. The operation PosLinear($a,b$) represents the usual Linear transformation with $a$ inputs and $b$ outputs, but considering only positive weights. To compare between having and not having uncertainty regularization, we use two different approximations of the standard deviation in the decoder: $\sigma_{\text{UR}}$ when performing meaningful uncertainty quantification, and $\sigma_{\text{ no UR}}$ otherwise.}
\label{appendix:tab:UQ_experiment_networks}
\end{table}

\begin{table}
    \centering
    \begin{tabular}{rllrr}
        \multicolumn{5}{c}{\textbf{Toy latent spaces} (Sec. \ref{sec:toy_latent_space_experiment})}\\
        \toprule
        \textbf{Distribution} & \textbf{Module} & \textbf{MLP} & \textbf{Seed for randomness} & $\beta$ \textbf{in} $\tilde{\sigma}_\beta$ \\
         \midrule
        \multirow{2}{*}{Normal} & $\mu$ & Linear(2,3) & \multirow{2}{*}{1} & \multirow{2}{*}{-2.5} \\
         & $\sigma$ & Linear(2,3), Softplus() & & \\
         \midrule
         Bernoulli & $p$ & Linear(2,15), Sigmoid() & 1 & -3.5 \\
         \midrule
         \multirow{2}{*}{Beta} & $\alpha$ & Linear(2,3), Softplus() & \multirow{2}{*}{1} & \multirow{2}{*}{-4.0} \\
         & $\beta$ & Linear(2,3), Softplus() & & \\
         \midrule
         Dirichlet & $\alpha$ & Linear(2,3), Softplus() & 17 & -4.0 \\
         \midrule
         Exponential & $\lambda$ & Linear(2,3), Softplus() & 17 & -4.0 \\
         \bottomrule \\[0.1cm]
    \end{tabular}
    \caption{This table describes the neural networks used for the experiment presented in Sec. \ref{sec:toy_latent_space_experiment}. Following the notation of PyTorch, Linear($a,b$) represents an MLP layer with $a$ input nodes and $b$ output nodes. In each of these networks, we implement the reweighting operation described in Sec. \ref{appendix:sec:UQ}, and we describe the $\beta$ hyperparameter present in the modified sigmoid function (Eq. (\ref{appendix:eq:translated_sigmoid})). This networks were not trained in any way, and they were initialized using the provided seed.}
    \label{appendix:tab:toy_experiment_networks}
\end{table}

\begin{table}
\centering
\begin{tabular}{@{}rl@{}}
\multicolumn{2}{c}{\textbf{Decoding to a von Mises-Fisher Distribution} (Sec \ref{sec:motion-capture}, \ref{sec:experiment_numerical_approx_metric}, \ref{sec:LAND_experiment})} \\
\toprule
\textbf{Module}  & \multicolumn{1}{c}{\textbf{MLP}}    \\ \midrule
\multicolumn{2}{c}{Encoder (Normal dist.)}  \\[0.1cm]
$\mu$ & Linear($3\times 26$, 90), Linear(90, 2) \\
$\sigma$ & Linear($3\times 26$, 90), Linear(90, 2), Softplus() \\
\midrule
\multicolumn{2}{c}{Decoder (vMF dist.)}  \\[0.1cm]
$\mu$ & Linear(2, 90), Linear(90, $3\times 26$), Linear($3\times 26$, $3\times 26$)  \\
$\kappa$ & Linear(2, 90), Linear(90, $3\times 26$), Linear($3\times 26$, $26$), Softplus()  \\
\midrule
Optimizer  & \multicolumn{1}{@{}l@{}}{Adam ($\alpha = 1 \times 10^{-3}$)}  \\
Batch size             & \multicolumn{1}{@{}l@{}}{16} \\
$\beta$ in $\tilde{\sigma}_\beta$             & \multicolumn{1}{@{}l@{}}{-5.5} \\
KL annealing & \multicolumn{1}{@{}l@{}}{0.01}\\
Extrapolation mechanism & \multicolumn{1}{@{}l@{}}{$\kappa \to 0.1$}\\
\bottomrule
\end{tabular}
\vspace{0.2cm}
\caption{This table shows the Variational Autoencoder used in our last two experiments (see Sec. \ref{sec:motion-capture}, \ref{sec:experiment_numerical_approx_metric}). Our motion capture data tracked 26 different bones, and thus we decode to a product of 26 different von Mises-Fisher distributions. }
\label{appendix:tab:vMF_experiments} 
\end{table}

\begin{table}
\centering
\begin{tabular}{@{}rl@{}}
\multicolumn{2}{c}{\textbf{Decoding to a Bernoulli Distribution} (Sec \ref{sec:movie-ratings})} \\
\toprule
\textbf{Module}  & \multicolumn{1}{c}{\textbf{MLP}}    \\ \midrule
\multicolumn{2}{c}{Encoder (Normal dist.)}  \\[0.1cm]
$\mu$ & Linear(60, 16), Tanh(), Linear(16, 16), Tanh(), Linear(16, 2) \\
$\sigma$ & Linear(60, 16), Tanh(), Linear(16, 16), Tanh(), Linear(16, 2), Softplus() \\
\midrule
\multicolumn{2}{c}{Decoder (Bernoulli dist.)}  \\[0.1cm]
$p$ & Linear(2, 16), Tanh(), Linear(16, 16), Tanh(),  Linear(16, 60), Sigmoid()  \\
\midrule
Optimizer  & \multicolumn{1}{@{}l@{}}{Adam ($\alpha = 1 \times 10^{-3}$, $\omega = 1 \times 10^{-7}$)}  \\
Batch size             & \multicolumn{1}{@{}l@{}}{256} \\
$\beta$ in $\tilde{\sigma}_\beta$             & \multicolumn{1}{@{}l@{}}{-3.0} \\
KL annealing & \multicolumn{1}{@{}l@{}}{0.01}\\
Extrapolation mechanism & \multicolumn{1}{@{}l@{}}{$p\to 1/2$} \\
\bottomrule
\end{tabular}
\vspace{0.2cm}
\caption{This table shows the Variational Autoencoder used in the movie rating experiement (see Sec. \ref{sec:movie-ratings}). The MovieLens 25M dataset has been preprocessed such that it is composed of 10000 users rating if they have seen some of 60 selected movies. We only select users that have seen more than two movies and less than 30 movies, to avoid outliers and aim for a more realistic scenario. We used the same extrapolation mechanism described in the toy experiments for the Bernoulli: having the probits be $1/2$ (see Sec. \ref{appendix:sec:UQ}).}
\label{appendix:tab:Bernoulli_experiments} 
\end{table}

In this section we describe, in detail, the models that we used for our experiments (see Sec.\ref{sec:experiments}). All the networks that we used are Multi-Layer Perceptrons implemented in PyTorch.

First, Table \ref{appendix:tab:UQ_experiment_networks} shows the Variational Autoencoder implemented for the experiment described in Sec. \ref{sec:compare_metrics_with_gaussians_generators_uq}. In the computations \textit{without} uncertainty regularization, we used a simpler model for the uncertainty quantification (namely, a single Linear layer, followed by a Softplus activation). For our second experiment involving a toy latent space, we also provide the implementation of the respective MLPs in Table \ref{appendix:tab:toy_experiment_networks}.
Finally, Table \ref{appendix:tab:vMF_experiments} and \ref{appendix:tab:Bernoulli_experiments} respectively represents the VAE trained for the experiments related to motion capture (Sec. \ref{sec:motion-capture}) and movie rating (Sec. \ref{sec:movie-ratings}). For the motion capture experiments, we are training a VAE that decodes to a von Mises Fisher distribution, and for the movie rating experiments, we decode to a Bernoulli distribution.

All of these VAEs were trained by maximizing the Evidence Lower Bound with different values for KL annealing which can be read from the different tables. For example, Table \ref{appendix:tab:vMF_experiments} shows that the KL annealing constant was chosen to be $0.01$.

\subsection{Metric approximation and KL by sampling}

When visualising our latent space as a statistical manifold, we can obtain a direct approximation of the metric using the KL-divergence between two close distributions (Proposition \ref{prop:approximating_metric}). We will show here, in simple cases, how our metric approximation compares to close-form expressions. 

In the following experiment, our statistical manifold is the parameter space of known distributions (Beta and Normal). Their Fisher-Rao matrices are well-known (Sec. \ref{appendix:FR:distributions}), and we approximate them by computing the KL-divergence of sampled distributions. We call $\b{M}_{t}$ the theoretical metric and $\b{M}_{a}$ the approximated metric, and we note $\varepsilon_{r} = \frac{\lVert \b{M}_t - \b{M}_a \rVert}{\lVert \b{M}_t\rVert}$ the relative error between the theoretical and approximated matrices. Here, $\lVert \cdot \rVert$ denotes the Frobenius norm. For the Normal distribution, we empirically obtain: $\varepsilon_{r} =  5.32 \cdot 10^{-4} \pm 9.63 \cdot 10^{-4}$, and for the Beta distribution, we have: $\varepsilon_{r} =  1.73 \cdot 10^{-5} \pm 1.17 \cdot 10^{-5}$.

\subsection{Computational complexity}

Proposition \ref{prop:FRapproxKL} shows the system of equations required to approximate the pullback metric in the latent space. Each KL operation requires 2 forward passes from the decoder to compute, so first we establish the lower bound on the time complexity of the decoder forward pass. Ignoring all activation function related operations, for an MLP with $H$ hidden layers, $N$-dimensional network output, $K$-dimensional hidden layer output and single $M$-dimensional vector input, this lower bound is:

\begin{align}
    \Omega \left( MK_1 + K_HN + \sum_{i=1}^{H-1} M_i M_{i+1} \right)
\end{align}

For each diagonal element $\b{M}_{ii}$ of the metric tensor we need to compute a single KL divergence, which will require two forward passes through the decoder network giving us a (lower bounded) time complexity of $\Omega \left[ 2 \left( MK_1 + K_HN + \sum_{i=1}^{H-1} M_i M_{i+1} \right) \right]$ for each element. For the off-diagonal elements we will need to compute the KL three times which corresponds to six forward passes through the decoder network. which yields a (lower bounded) time complexity $\Omega \left[ 6 \left( MK_1 + K_HN + \sum_{i=1}^{H-1} M_i M_{i+1} \right) \right]$ per element.

\subsection{Information for the movie preferences experiment}
\begin{figure}[h]
    \centering
  \includegraphics[width=0.4\linewidth]{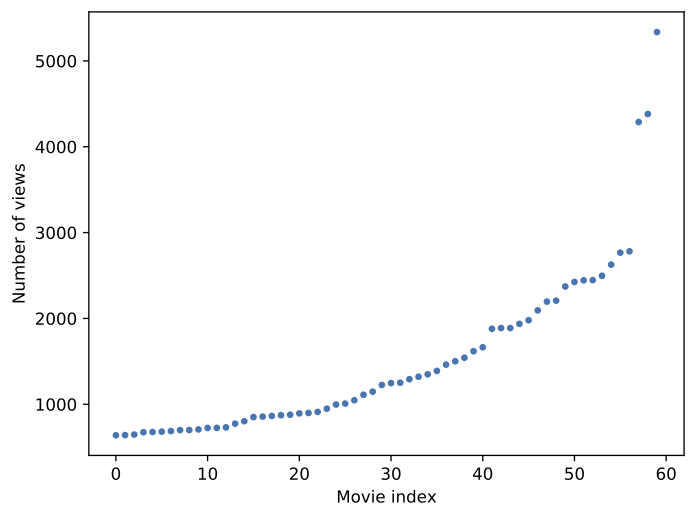}
  ~
  \includegraphics[width=0.4\linewidth]{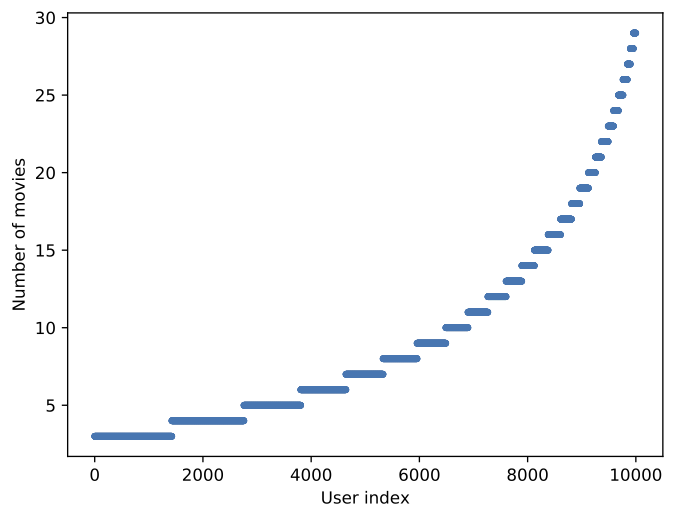}
  \caption{The numbers of views for the movies and the users.}
  \label{app:fig:movies_users_numbers}
\end{figure}

For this experiment we used the MovieLens 25M dataset (\url{https:// grouplens.org/datasets/movielens/25m/}). Each cell of the data matrix represents the rating of a user (row) from 1 to 5 for the corresponding movie (column). In order to fit a Bernouli VAE we considered the matrix as binary i.e. if a user has seen a movie (1) or not (0). We then selected the 60 most popular movies, as well as, 10000 users who have seen between 2 and 30 of these movies. We also verified that all the movies have been seen from at least 600 users. In this way we reduced the size of the dataset, obtaining a realistic scenario where: 1) some movies are more popular than the others, and 2) we do not include users that have seen 0 or almost all the movies. We show in Fig.~\ref{app:fig:movies_users_numbers} the number of views for each movie and the number of movies each user has seen. In Table~\ref{appendix:tab:Bernoulli_experiments} we present the details for the Bernouli VAE.

\subsection{Information for fitting the LAND model}

The locally adaptive normal distribution (LAND) \citep{arvanitidis:nips:2016} is the extension of the normal distribution on Riemannian manifolds learned from data. \citet{pennec:jmiv:2006} first derived this distribution on predefined manifolds as the sphere and also showed that it is the maximum entropy distribution given a mean and a precision matrix. The flexibility of this probability density relies on the shortest paths. However, the computational demand to fit this model is relatively high, especially in our case, since we need to use an approximation scheme to find the shortest paths.

In particular, we compute the \emph{logarithmic map} $\b{v}=\mathrm{Log}_\x(\b{y})$ by first finding the shortest path between $\x$ and $\b{y}$, and then, rescaling the initial velocity as $\b{v} = \frac{\dot{c}(0)}{||\dot{c}(0)||} \mathrm{Length}(c)$, which ensures that $||\b{v}|| = \mathrm{Length}(c)$. In addition, for the estimation of the normalization constant we use the \emph{exponential map} $\mathrm{Exp}_\x(\b{v})=c_\b{v}(t)$, which is the inverse operator that generates the shortest path with $c(1)=\b{y}$ taking the rescaled initial velocity $\b{v}$ as input. Also, we should be able to evaluate the metric. While the logarithmic map can be approximated using our approach (Section~\ref{sec:geodesic_solver_details}), for the exponential map we need to solve the ODEs system \eqref{eq:ode_system} as an initial value problem (IVP). Note that we fit the LAND using gradient descent, which implies that the computation of these operators is the main computational bottleneck.

We provided a method in Proposition~\ref{prop:approximating_metric}, which enables us to approximate the pullback metric in the latent space of a generative model using the corresponding KL divergence. Even if this is a sensible approach, in practice, the computational cost is relatively high as we might need to estimate the KL using Monte Carlo. For example, this is the case when the likelihood is the von Mises-Fisher. This further implies that fitting the LAND using this approach is prohibited due to the computational cost. Especially, since we need to evaluate many times the metric and its derivative for the computation of each exponential map. Hence, in order to fit the LAND efficiently, we used the following approximation based on \citet{hauberg:nips:2012}.

First we construct a uniformly spaced grid in the latent space. Then, we evaluate the metric using Proposition~\ref{prop:approximating_metric} for each point on the grid getting a set $\{\z_s, \b{M}_s\}_{s=1}^S$ of metric tensors. Thus, we can estimate the metric at any point $\z$ as
\begin{equation}
    \b{M}(\z) = \sum_{s=1}^S \widetilde{w}_s(\z) \b{M}_s, ~~ \text{with} ~~ \widetilde{w}_s(\z) = \frac{w_s(\z)}{\sum_{j=1}^S w_s(\z)} ~~ \text{and} ~~ w_s(\z) = \exp\left(-\frac{ ||\z_s - \z||^2}{2\sigma^2}\right)
\end{equation}
where $\sigma >0$ the bandwidth parameter. This is by definition a Riemannian metric as a weighted sum of Riemannian metrics with a smooth weighting function. In this way, we can approximate the pullback of the Fisher-Rao metric in the latent space $\Z$ in order to perform the necessary computations more efficiently.

\clearpage
\vfill

\end{document}